\def\year{2022}\relax
\documentclass[letterpaper]{article} 
\usepackage{aaai22}  
\usepackage{times}  
\usepackage{helvet}  
\usepackage{courier}  
\usepackage[hyphens]{url}  
\usepackage{graphicx} 
\usepackage{natbib}  
\usepackage{caption} 
\DeclareCaptionStyle{ruled}{labelfont=normalfont,labelsep=colon,strut=off} 
\usepackage{algorithm}
\usepackage{algorithmic}

\usepackage{newfloat}
\usepackage{listings}
\floatstyle{ruled}
\newfloat{listing}{tb}{lst}{}
\floatname{listing}{Listing}
\title{Gender and Racial Stereotype Detection in Legal Opinion Word Embeddings}
\author{
    Sean Matthews\textsuperscript{\rm 1}\equalcontrib,
    John Hudzina\textsuperscript{\rm 1}\equalcontrib,
    Dawn Sepehr\textsuperscript{\rm 2}
}
\affiliations{
    \textsuperscript{\rm 1} Thomson Reuters Labs, Eagan, Minnesota, USA\\
    \textsuperscript{\rm 2} Thomson Reuters Labs, Toronto, Canada\\
    sean.matthews@thomsonreuters.com, john.hudzina@thomsonreuters.com, dawn.sepehr@thomsonreuters.com
}

\usepackage{tcolorbox}
\usepackage{booktabs}
\usepackage{footmisc}
\usepackage{cleveref}

\begin{document}

\maketitle


\begin{abstract}
Studies have shown that some Natural Language Processing (NLP) systems encode and replicate harmful biases with potential adverse ethical effects in our society.
In this article, we propose an approach for identifying gender and racial stereotypes in word embeddings trained on judicial opinions from U.S. case law.
Embeddings containing stereotype information may cause harm when used by downstream systems for classification, information extraction, question answering, or other machine learning systems used to build legal research tools.
We first explain how previously proposed methods for identifying these biases are not well suited for use with word embeddings trained on legal opinion text.
We then propose a domain adapted method for identifying gender and racial biases in the legal domain.
Our analyses using these methods suggest that racial and gender biases are encoded into word embeddings trained on legal opinions.
These biases are not mitigated by exclusion of historical data, and appear across multiple large topical areas of the law.
Implications for downstream systems that use legal opinion word embeddings and suggestions for potential mitigation strategies based on our observations are also discussed.
\end{abstract}

\section{Introduction}
Recent developments in the field of Artificial Intelligence (AI) have transformed the way data is prepared and turned into information for interpretation in different domains spanning from social media to legal documents. 
These advancements have predominantly paved the way for creating more accurate predictive models, however, multiple research studies have shown that these systems are not without fault and have inadvertently perpetuated some harmful biases and stereotypes present in society by encoding and replicating patterns of bias present in the data upon which they are trained.
Examples of such faulty systems include racial bias detected in hate speech predictive models for social media posts \cite{mozafari2020hatespeech}, unequal distribution of health care resources across racial groups due to incorrectly identifying patients who need significant healthcare \cite{obermeyer2019dissecting}, displaying fewer Science, Technology, Engineering, and Mathematics (STEM) job advertisements to women compared to men \cite{lambrecht2019algorithmic}, and racial disparities demonstrated in recidivism risk prediction algorithms \cite{dieterich2016compas}. 

While these tools are becoming more and more integrated in our societies and extend great benefits when deployed properly, they also pose high risks of imposing unfair life changing decision making upon minorities and more vulnerable communities.
This may be particularly important when developing technologies used within the legal system due to the significant impacts the legal system in general has on individuals, businesses, government entities, and many other aspects of society.
Deploying predictive technologies based on biased models into contexts where they are used by individuals interacting with the legal system at various levels could potentially result in a broad array of harmful effects in society including decreased quality of legal representation, increased costs associated with litigation, or even increased likelihood or duration of incarceration for individuals belonging to groups affected by these biases.
Hence, it is imperative that we take steps towards identifying, mitigating, and ultimately eliminating these undesired effects in legal technologies relying on predictive systems.

We would like to emphasize that historical bias is not the only form of bias that can be found in AI systems: representation, measurement, aggregation, evaluation, and deployment biases have also been identified at different stages of developing an AI system \cite{suresh2020framework}.
In this article, however, we mainly focus on revealing historical and representational bias found in word embeddings trained on judicial opinions from U.S. case law and the distinct challenges that arise when developing predictive models in the legal domain.

\subsection{Bias in Word Embeddings}
Word embedding approaches such as word2vec \cite{mikolov2013efficient, mikolov2013distributed}, GloVe \cite{pennington-etal-2014-glove}, etc., represent words in an $n$-dimensional space by encoding contextual co-occurrence statistics for words occurring in large text corpora. 
Since these associations are obtained from compiling large historical corpora, different types of biases that already exist in these texts will inevitably plague the word representations if appropriate considerations are not anticipated. 
Multiple previous studies have investigated and shown the presence of these biases in the form of either benign or neutral effects such as associating flowers with pleasant words vs. associating weapons with unpleasant words, or detrimental effects by encoding discrimination based on protected categories such as race, gender, social status, etc. \cite{bolukbasi_man_2016, caliskan_semantics_2017}. 

Different methodologies have been proposed to identify, visualize, and mitigate these effects.
One prominent approach draws inspiration from a method originally developed in the field of social psychology to measure implicit bias in humans.
The Implicit Association Test (IAT) measures the differential response times of human participants while categorizing sets of target words (e.g., flower and insect names) and attribute terms (e.g., pleasant or unpleasant) when they are paired in stereotypical (e.g. flower-pleasant) or counter-stereotypical (e.g. flower-unpleasant) configurations \cite{greenwald1998measuring}. 
Both the stimuli used in the IAT and the general strategy of detecting bias through differential association strength have been adapted to develop bias detection strategies to measure bias encoded in word embeddings such as the Word Embedding Association Test \cite[WEAT;][]{caliskan_semantics_2017}. 
The WEAT measures this difference in association strength between two groups by calculating the similarity of the embeddings in a set of target words used as a proxy for group membership (e.g., common female given names) with the embeddings in two sets of attribute words (e.g., pleasant and unpleasant terms) and computing the difference between these similarities, then comparing the difference in these association strengths to the same difference score calculated for a second target group (e.g., common male given names). 
Using methods based on this test, Rice et. al found evidence for  racial biases being encoded in word embeddings trained on legal texts such as appellate court opinions from US state and federal courts \cite{rice_racial_2019}.

\subsection{Legal Word Embedding Issues}
As mentioned in the previous section, the social impact of encoded bias in word embeddings is becoming more significant in the legal domain which has direct implications on many aspects of our society as legal technologies using these types of representations gain greater adoption.
In this section, we review some of the challenges that arise when working with legal text corpora and in the subsequent sections we present our solution for addressing these issues. 

\subsubsection{Names in Legal Text:}
Although the WEAT racial and gender stereotype tests relied on given names \cite{caliskan_semantics_2017}, legal opinions construct more formal sentences than the wikipedia and news articles used to train the publically available GloVe embeddings \cite{pennington-etal-2014-glove}.  
For example, Figure~\ref{fig:my_label} demonstrates the co-referencing of a natural person in legal opinions.  Note that Gerald Bostock's given name is only referenced once.  In most cases, the natural person's full name is typically referenced first followed by the surname and/or pronouns thereafter.  If a legal system applied the given name tests only, then bias encoded in surnames and gendered pronoun embeddings would be missed.    

\begin{figure}[tb]
    \centering
    \begin{tcolorbox}
    Excerpt from Bostock v. Clayton County: \\
        
    \textbf{Gerald Bostock} worked for Clayton County, Georgia, as a child welfare advocate. Under \textbf{his} leadership, the county won national awards for its work. After a decade with the county, \textbf{Mr. Bostock} began participating in a gay recreational softball league. Not long after that, influential members of the community allegedly made disparaging comments about \textbf{Mr. Bostock's} sexual orientation and participation in the league. Soon, \textbf{he} was fired for conduct “unbecoming” a county employee.
    \end{tcolorbox}
    \caption{Legal Opinion Co-referencing Example}
    \label{fig:my_label}
\end{figure}

Another concern specific to the legal domain is that legal opinions potentially may introduce gender-occupational stereotypes because they typically state a judge's full name and judicial title.  
Historically, women only account for 12.3\% of federal Title III judicial appointments \cite{fjcexport}. 
Given that \citeauthor{caliskan_semantics_2017} \citeyearpar{caliskan_semantics_2017} found significant gender-occupation bias in non-legal text and the historical imbalance of female judges, embeddings built upon legal opinions potentially perpetuate this specific stereotype. 

\subsubsection{Positive \& Negative Sentiment:}

Whereas WEAT evaluated sentiment in a generalized modern web corpus, the legal opinions contain historical domain specific terminology.
The WEAT study evaluated several tests measuring positive and negative sentiment for various target groups.  
These tests are from the IAT with very small vocabularies as required due to fatigue effects in human participants \cite{caliskan_semantics_2017}. 
The sentiment-based test must be adjusted for legal opinions because the general vocabularies used to describe positive and negative sentiment do not align with how positive and negative sentiment is expressed in judicial opinions \cite{rice_corpus-based_2021}.

\subsubsection{Legal Outcomes:} 

While the IAT tests mainly focus on negative or positive attributes, legal outcome extraction provides a greater risk of harm to protected classes than sentiment analysis.  Courts document legal outcomes in docket entries, orders, judgements, and/or opinions.  Litigation analytics extract legal outcomes for these free text sources because many jurisdictions do not record outcomes in a structured form at a party level \cite{vacek-etal-2019-litigation}.  If the word embeddings influence outcome extraction based on a party's gender or race, then embedding-based analytics may amplify racial and gender bias by causing parties to settle for something other than their case's merits.   

\subsection{Contributions}
As discussed previously, word embeddings are used in many practical NLP systems which operate on legal language.
In this article, we propose an approach for identifying racial and gender biases encoded in word embeddings that are created using the text of legal opinions. 
This approach addresses multiple issues specific to legal language that have not been addressed in the previous work.
These challenges deal with idiomatic phrases as well as specific considerations for adapting the WEAT tests to legal language for detection of bias. 
We also investigate how these biases have changed over time as well as their strengths in different topical areas of the law.

\section{Proposed Approach}  \label{sec:approach}
In this section we describe the main approach proposed for identifying bias in word embeddings created based on legal opinions. 
We first describe the legal corpus under study and the required data preparation.
Next, we briefly discuss the Word Embedding Association Test (WEAT).
Finally, we state how we addressed the challenges identified in the previous section with domain adapted tests.

\subsection{Opinion Preparation \& Embedding Construction}

\begin{figure}[b]
    \centering
    \includegraphics[scale=.57]{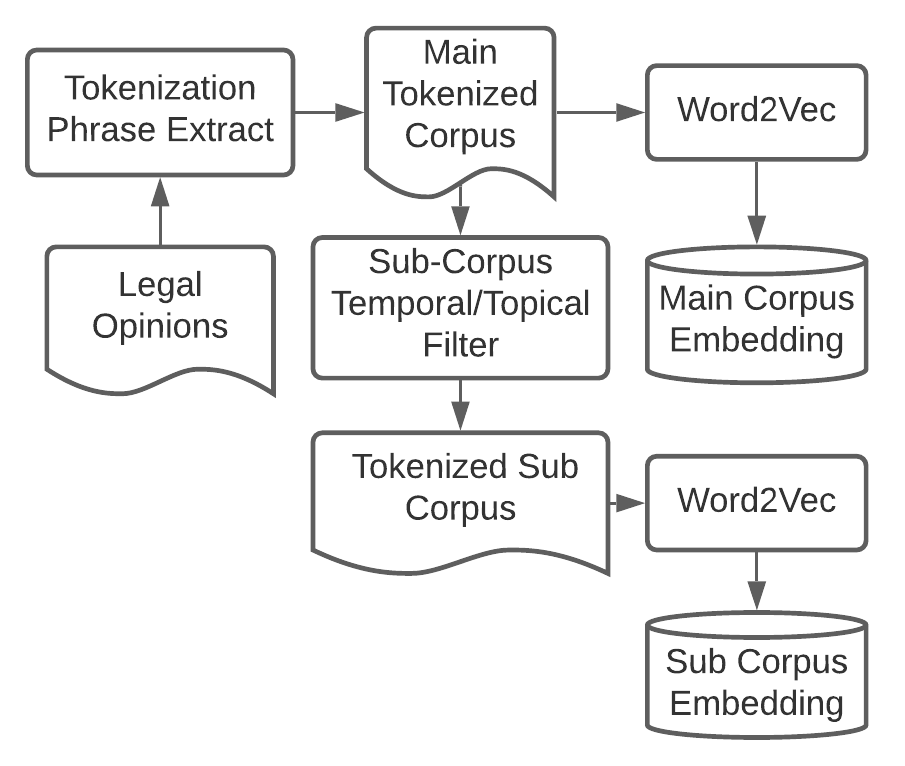}
    \caption{Corpus Prep \& Embedding Generation}
    \label{fig:prep}
\end{figure}

For our experiments, we examined embeddings created from a large corpus of U.S. legal opinions. 
The corpus includes over 12 million opinions from 1,949 current and historic jurisdictions dating back to 1650.  
The corpus size contains 10x more opinions than a previous legal opinion bias study \cite{rice_racial_2019}.  
The main corpus includes U.S. federal, state, and territorial courts with the notable exception of tribal courts.  
The tribal court opinions are handled as a supplemental corpus from the source system.  
To generate the embeddings for the full corpus, topical sub-corpora and historical sub-corpora, we follow the process in Figure~\ref{fig:prep}. 

\subsubsection{Idiomatic Phrase Extraction:}
Prior to generating the embeddings, we extracted idiomatic phrases.
Non-contextual word embeddings assume phrase meanings are composed from representations of individual words.  
However, this composability assumption does not always hold true for legal jargon and idiomatic phrases.  
For example, the Latin phrase \emph{pro hac vice} means "for this time only" and does not have the same semantic meaning as the individual words \emph{pro, hac, \& vice}.  

Although contextual embeddings handle this issue by representing the relative relationships between words, non-contextual embeddings only represent idiomatic phrases as single tokens \cite{mikolov2013distributed}. 
To avoid overly large n-gram dictionaries, the phrase extractor only combines tokens that commonly appear together. 
Our phrase extractor used a Normalized Point-wise Mutual Information (NPMI) score to select the n-grams to add in the dictionary \cite{bouma_normalized_2009}.
NPMI scores range from -1 (never co-occurs) to 1 (always co-occurs), with 0 meaning the tokens are completely independent.  
We ran two passes of the phrase extractor that selected phrases with a minimum NPMI score of 0.5. 

\subsubsection{Embedding Training:}

Once the phrase extraction was completed, we trained the embeddings against the complete corpus, as well as sub-corpora for temporal cutoff dates, and divided by topic (see section "Experimental Results"). 
Each embedding followed the same training procedure using a skip-gram word2vec model.
For all embeddings, the hyper-parameters included a 300 dimension vector size, a minimum term frequency of 30, a $10^{-4}$ sampling threshold, a learning rate of 0.05, a window size of 10, and 10 negative samples.

\subsection{Word Embedding Association Test}

Our experiments use \citeauthor{caliskan_semantics_2017} \citeyearpar{caliskan_semantics_2017} original word lists applied to the legal opinion embeddings, as well as tests based on domain specific and expanded word lists. 
For each test, which includes both the target $X$ and $Y$ word lists, and the attribute $A$ and $B$ word lists, we calculate the effect size (Cohen's $d$).  
We also calculate the standard error by sub-sampling the word lists with a simple bootstrapping procedure.

\subsection{Domain Adaptation}

Once the embeddings were trained, we extended the \citeauthor{caliskan_semantics_2017} tests with new domain specific tests.  
This section details the methodologies used to generate legal specific target and attributes terms for the new tests.
These updates include new attribute lists:
\begin{itemize}
    \item Positive vs. Negative Legal 
    \item Legal (Motion) Outcome
    \item Expanded Career vs. Family
\end{itemize}
The domain updates also include new target lists:
\begin{itemize}
    \item Surnames by Race
    \item Male vs. Female Terms
    \item Judge Given Names
\end{itemize}

\subsubsection{Positive vs. Negative Legal:}

In order to generate a legal specific sentiment vocabulary, we implemented a minimally supervised approach developed by \citeauthor{rice_corpus-based_2021} \citeyearpar{rice_corpus-based_2021}. 
This work provided a legal specific list of positive, $V_p$, and negative, $V_n$, seed terms \cite{dvn_2019} and a method for expanding the term sets. 
We then generated the expanded list based on the legal opinion corpus embeddings. 
The expanded positive valence terms were found by a cosine similarity search on the vector $\sum \vec{V}_p - \sum \vec{V}_n$.  
Conversely, The negative valence terms were found on a cosine similarity search on the vector $\sum \vec{V}_n - \sum \vec{V}_p$.
The expanded term lists were then manually reviewed to exclude any terms with obvious race or gender associations (e.g. "gentlemanly")\footnote{\label{fn-supp-mat}See the Supplementary Material for excluded terms: https://arxiv.org/abs/2203.13369}.

\subsubsection{Legal (Motion) Outcome:}

Trial motions are discussed and reviewed within legal opinions.  
This text typically includes the party's surname, motion type, and disposition \cite{vacek-etal-2019-litigation}.  
The manually created "Grant vs. Deny" attribute lists capture the positive and negative outcomes for a given motion.

\subsubsection{Expanded Career vs. Family List:}

The expanded career list employed the same minimally supervised approach as the positive and negative legal attribute list expansion. Instead of finding terms along a positive/negative dimension of interest, we extracted terms along a career versus family axis in the embedding space.
The seed lists contained the career and family terms from \citeauthor{caliskan_semantics_2017}~\citeyearpar{caliskan_semantics_2017}.

\subsubsection{Surnames by Race:} \label{name-list}

As noted in the introduction, court documents reference parties by their surnames throughout the document.  
For the racial stereotype experiments, we used the surnames list from the 2010 U.S. decennial census as a proxy for race similar to medical outcome studies \cite{kallus_assessing_2020}.  
The census provides an estimated percentage of each race by surname.  
We sampled names from the list with over a 90\% probability for a given race.  

Although the U.S. Census provided a list of surnames, the referenced name is not guaranteed to refer to a natural person.
Instead, the name may reference a legal person's (i.e., corporation) name, a place name, or a thing.  
To reduce the potential name overlap with common words, we employed three methods to either reduce and or eliminate multi-sense words from the surname list: 
\begin{itemize}
    \item Title cased the surnames to target proper nouns. 
    \item Idiomatic phrase extraction to exclude non-person names like the \emph{State of Washington} \cite{mikolov2013distributed}.
    \item Centroid-based filtering to remove multi-sense words. 
\end{itemize}

The centroid-based filter removes candidate surnames based on the following procedure developed for WEAT \cite{caliskan_semantics_2017}.  
We computed a centroid vector based on the embedding vectors for all surnames in the U.S. 2010 Census and then computed the cosine similarity for each surname relative to the centroid.  
Finally, we removed 20\% of the least similar names.  
Once the filter was applied, we created the name lists for each test.  
While our target sample size was 200 surnames with at least 300 opinions per racial group, those criteria were not achievable for all races.  
Specifically, Native American and Alaskan names were proportionally underrepresented in the main corpus because fewer tribal court jurisdictions publish to and or are collected by the source system compared to State and Federal jurisdictions.
Table~\ref{tab:surnames} shows the sample sizes for each group.  
We adjusted the sample size for each test pair of surnames based on the smallest sample size in the pair.
\begin{table}[tb]
    \centering
    \begin{tabular}{l c c }
    \toprule
    Group & Sample Size & Min. Cases\\
    \midrule
    European & 46 - 200 & 300 \\
    African American & 164 & 300 \\
    Hispanic & 200 & 300 \\
    Asian Pacific Islander & 200 & 300 \\
    Native American / Alaskan & 46 & 30 \\
    \bottomrule
    \end{tabular}
     \caption{Surname Lists by Race}
     \label{tab:surnames}
\end{table}

\subsubsection{Male vs. Female Terms:}

A similar problem exists with the gender tests based on given names since individuals are often referred to primarily by their surnames throughout legal opinions.
To address this issue, we created a list of gendered pronouns and common gendered nouns (e.g. man/woman) for use in gender bias WEATs.

\subsubsection{Judge Given Name List:} \label{judgename-list}

In addition to the gendered first name list created by \citeauthor{caliskan_semantics_2017}~\citeyearpar{caliskan_semantics_2017}, we generated a gendered first name list based on judicial biographical data exported from the free law project's court listener \cite{judge_db}.
The biographical information included both race and gender for both State and Federal Judges.  
We calculate the percentages of female and male genders for each first name.  
For each gendered list we select names that occur at least 90\% of the time for that gender.

As with the surnames, some first names might overlap with place names, corporations, or other concepts.  
For example, Virginia might represent a Judge's name or a State.  
As with the surname we employed the following procedures:
\begin{itemize}
    \item Title cased the first name to target proper nouns. 
    \item Idiomatic phrase extraction to exclude non-person names, like the \emph{Commonwealth of Virginia}.
    \item Centroid-base filtering to remove multi-sense words. 
\end{itemize}

\section{Experimental Results}
\label{section:exp_result}

\subsection{Legal Opinion Corpus}

Before discussing the results for the legally adapted tests, we evaluate the opinion-based embedding using the \citeauthor{caliskan_semantics_2017} \citeyearpar{caliskan_semantics_2017} tests.  
Table \ref{tab:baseline} shows the baseline results\footnote{The Cohen's $d$ effect size ranges from -2.0 to 2.0 with $\pm0.5$ representing a medium effect}.  
While the legal opinion embeddings show a smaller effect size for the flower/insect control test than Caliskan, the opinion flower/insect control still exhibits a large effect size. 
In addition, the instrument/weapons control displays equivalent effect sizes between Caliskan's Common Crawl embeddings and the legal opinion embeddings.

Similar to the baseline tests, the racial and gender stereotype tests show a strong effect size.  
Note that Table~\ref{tab:baseline} uses the exact same target and attributes as \citeauthor{caliskan_semantics_2017} \citeyearpar{caliskan_semantics_2017}.
These tests use first names as the targets and non-legal terms as the attributes.  
Yet, we still see a moderate to strong effect for sentiment.  
The gender specific test replicated the occupational bias seen in past studies.  

\begin{table}[tb]
    \centering
    \begin{tabular}{l c c}
    \toprule
    Test & $d_C$ & $d_L$ \\
    \midrule
    Flowers/Insects \\
    \hspace{0.4cm} Pleasant vs Unpleasant & 1.50 & 0.97 \\
    Instruments/Weapons \\
    \hspace{0.4cm} Pleasant vs Unpleasant & 1.53 & 1.55 \\
    Eur. / African American names \\
    \hspace{0.4cm} Pleasant3 vs Unpleasant3 & 1.41 & 0.88 \\
    Male vs Female \\
    \hspace{0.4cm} Career vs Family & 1.81 & 1.75 \\
    \bottomrule
    \end{tabular}
    \caption{Cohen's effect size ($d$) comparison between Common Crawl GloVe ($d_C$) and Legal Opinion Word2Vec ($d_L$) }
    \label{tab:baseline}
\end{table}

\subsubsection{Gender Effects:}

While the effect sizes were comparable between the Common Crawl corpus and the legal corpus, the legal specific gender tests show some differences.  
Table~\ref{tab:gender} includes the original Caliskan and the new Legal attributes.  
The "Grant vs. Deny" tests all show a medium female negative bias for legal outcome.  
In comparison, the "Pleasant vs. Unpleasant" shows a positive female bias.  
In essence, positive sentiment does not necessarily relate to a positive outcome. 

\begin{table}[tb]
\centering
    \begin{tabular}{l c c}
    \toprule
    Test & $d$ & error \\
    \midrule
    Male vs Female Terms \\
    \hspace{0.4cm} Pleasant vs Unpleasant & -0.197 & 0.009 \\
    \hspace{0.4cm} Positive vs. Negative Legal & 0.089 & 0.007 \\
    \hspace{0.4cm} Grant vs Deny & 0.457 & 0.008 \\
    Male vs. Female Names (Judges) \\
    \hspace{0.4cm} Pleasant vs Unpleasant & -0.495 & 0.008 \\
    \hspace{0.4cm} Positive vs. Negative Legal & -0.254 & 0.003 \\
    \hspace{0.4cm} Grant vs Deny & 0.603 & 0.007 \\
    Male vs. Female Names (Caliskan) \\
    \hspace{0.4cm} Pleasant vs Unpleasant & 0.208 & 0.013 \\
    \hspace{0.4cm} Positive vs. Negative Legal & -0.198 & 0.003 \\
    \hspace{0.4cm} Grant vs Deny & 0.506 & 0.009 \\
    \bottomrule
    \end{tabular}
    \caption{Cohen's effect size ($d$) for gender specific test on the Legal Opinion Corpus}
    \label{tab:gender}
\end{table}

\subsubsection{Racial Effects:}
\begin{figure}[tb]
    \centering
    \includegraphics[scale=.36]{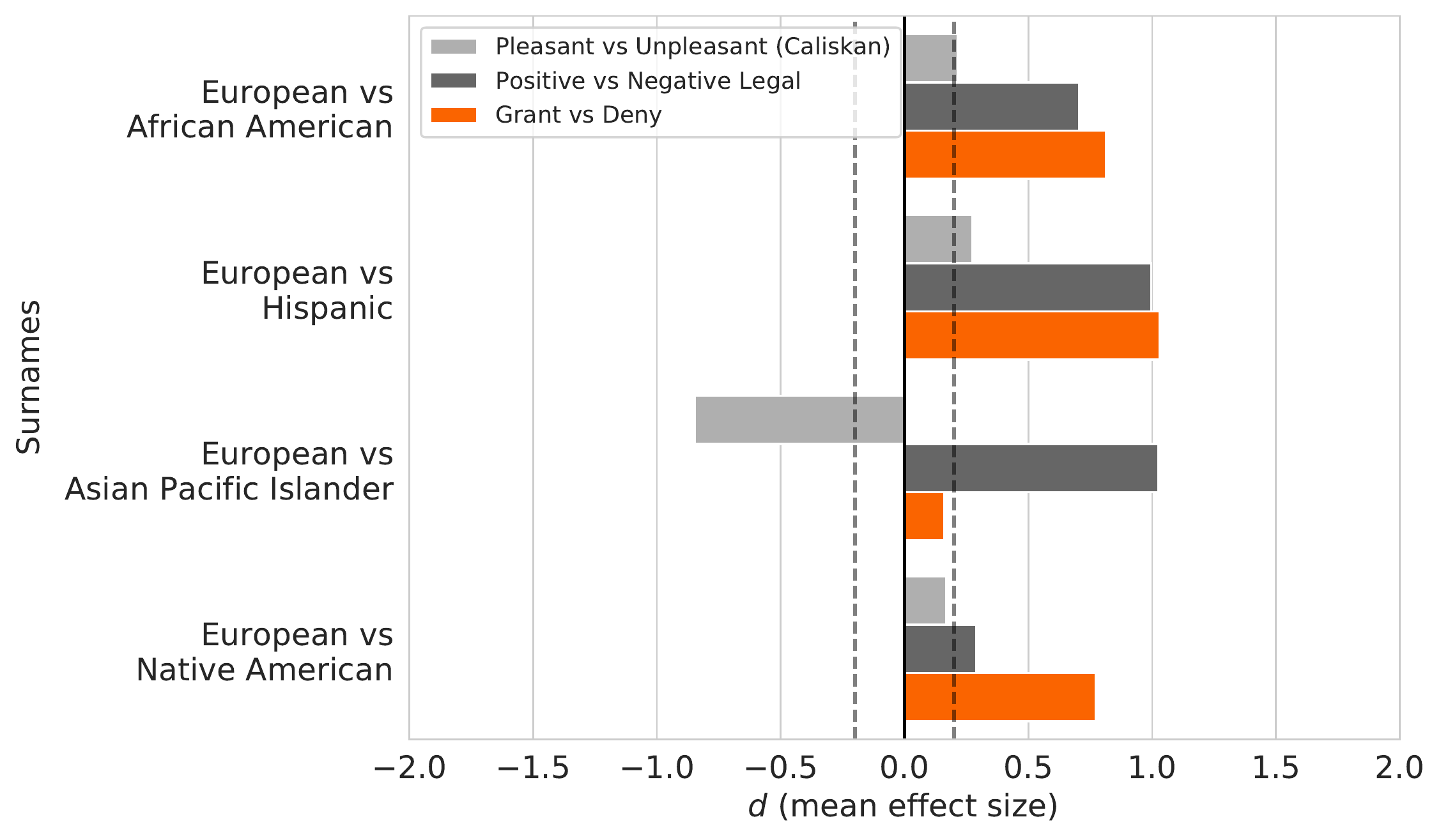}
    \caption{Surname WEAT Cohen's effect sizes}
    \label{fig:surname}
\end{figure}

As with the gender tests, we create both legal specific target and attribute word lists.  
Figure~\ref{fig:surname} shows the results for the surname-based racial bias experiments.  
The surname tests demonstrate a large difference in effect size between the "Pleasant vs. Unpleasant" sentiment and the Legal attributes.  
While the Hispanic surname tests only show a small negative effect for general sentiment, both legal specific tests showed a large negative effect.  

The Asian Pacific Islander results provided an even greater disparity in results than the Hispanic surname test.  Although the "Pleasant vs. Unpleasant" test showed a large positive bias for Asian Pacific Islanders, the "Positive vs. Negative Legal" test showed a large negative bias for Asian Pacific Islanders.  
In essence, positive stereotypes do not necessarily translate to group fairness in a legal context.

\subsection{Temporal Effects}
Given that the opinion corpus used to train the word embeddings contains opinions dating back to 1650, one possibility is that the observed gender and racial biases are driven by the inclusion of opinions from time periods where these biases were even more explicit and prevalent than in modern society. 
Our interest in the temporal component of these biases is primarily focused on representational harms that could be caused by NLP systems that may use representations similar to these in legal technology applications rather than in a historical analysis measuring the amount of bias present in any given time period.

\begin{figure}[tb]
    \centering
    \includegraphics[scale=.33]{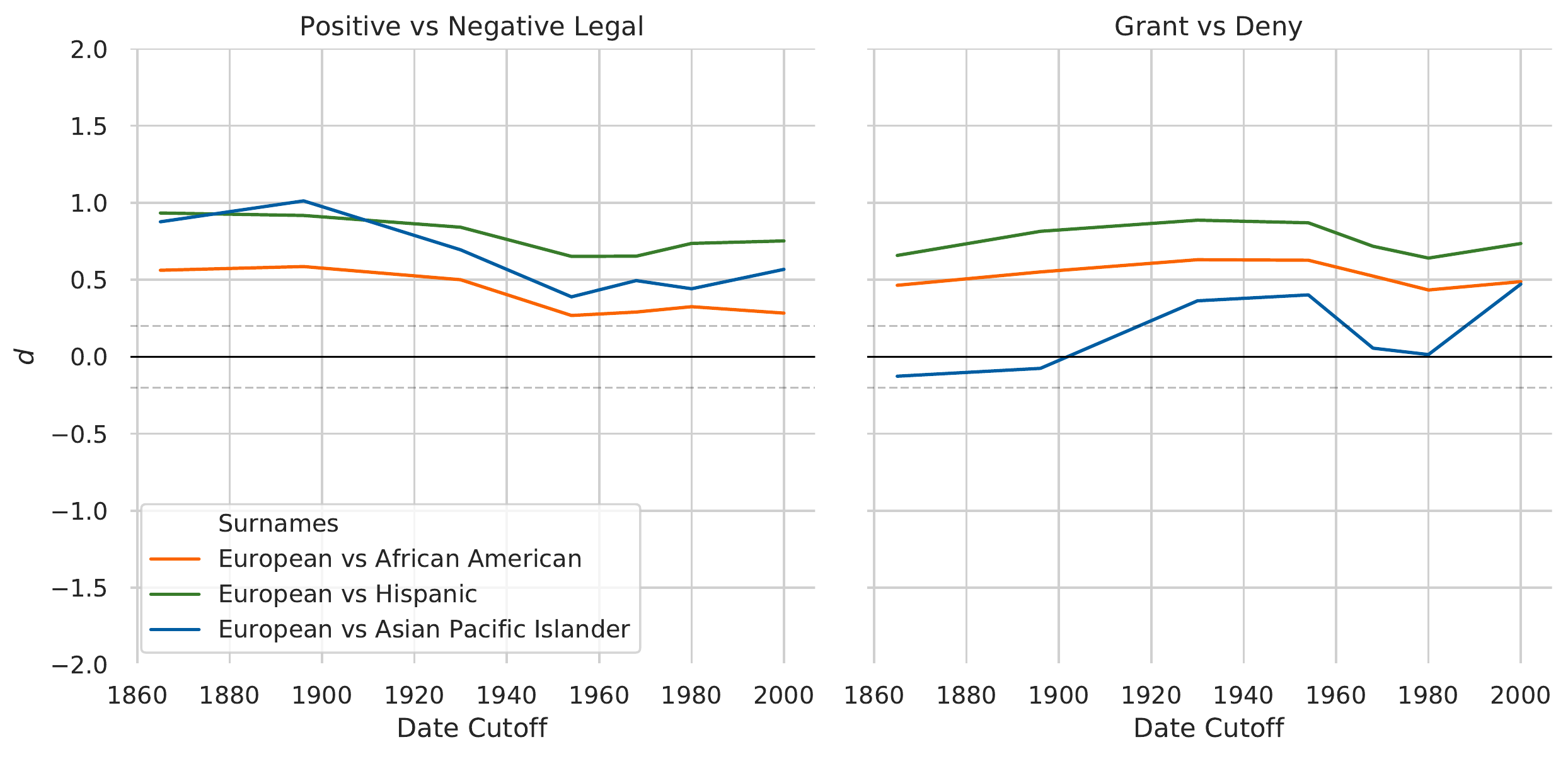}
    \caption{Temporal WEAT scores for race targets (surnames) and legal attributes}
    \label{fig:race-historical}
\end{figure}

\begin{figure}[tb]
    \centering
    \includegraphics[scale=.33]{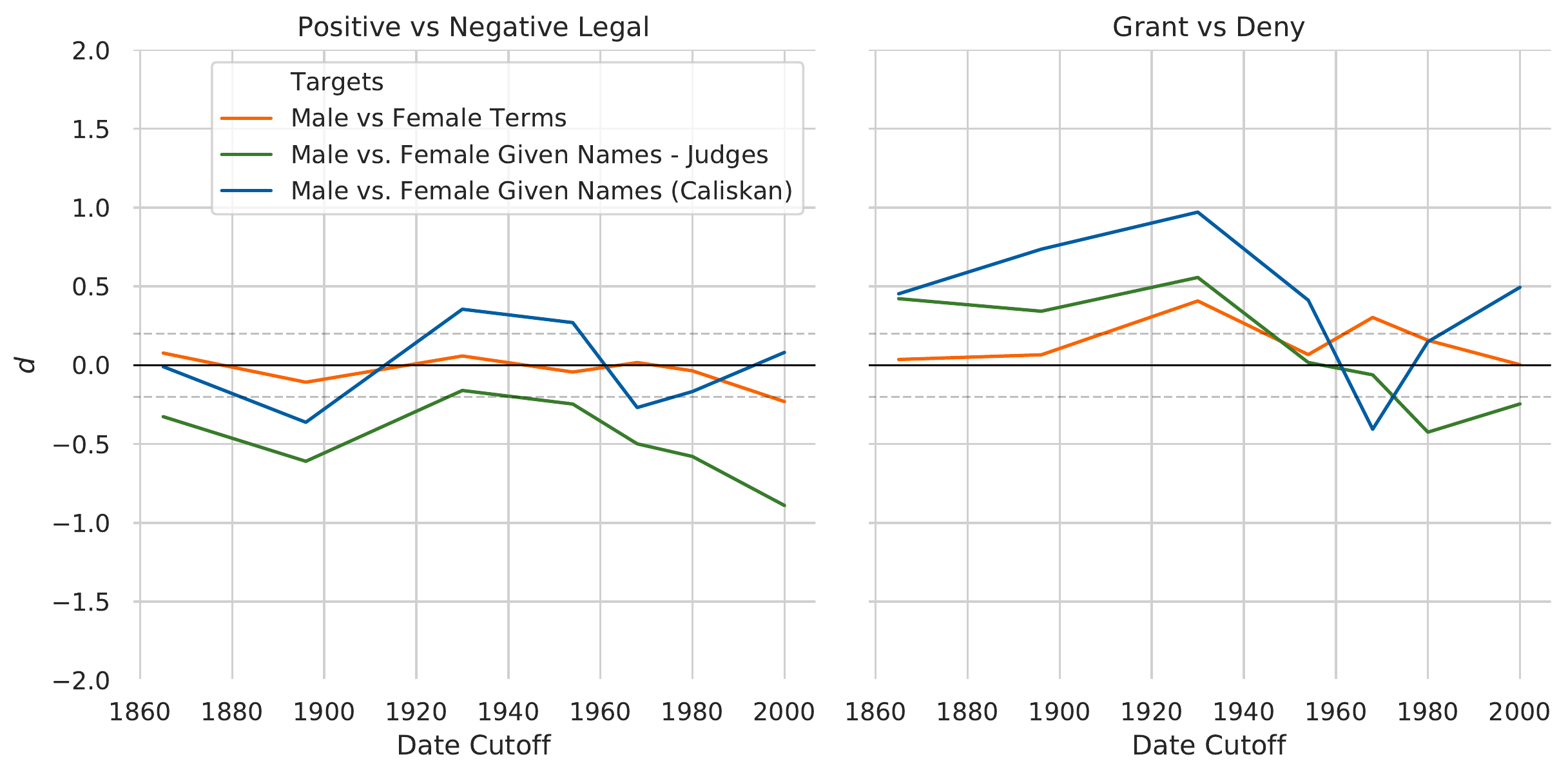}
    \caption{Temporal WEAT scores for gender-related targets and legal attributes}
    \label{fig:gender-historical-posneg}
\end{figure}

To investigate the effect of inclusion of historical opinions on the biases encoded in these representations, we trained word embeddings on temporal subsets of the corpora. 
These subsets were created by always including modern opinions, but varying the year of cutoffs for inclusion of historical data to incorporate older opinions into the corpus.  
The year cutoffs we selected were as follows: 2000-2020 (last 20 years), 1980-2020 (last 40 years), 1968-2020 (Post Civil Rights Act), 1954-2020 (Post Brown v. Board of Education), 1930-2020 (Post Great Depression), 1896-2020 (Post Plessy v. Ferguson), and 1865-2020 (Post Civil War).  
We then applied the legal-adapted WEAT analyses previously described to the embeddings generated for each temporal cutoff.
The results of these analyses are shown in Figures~\ref{fig:race-historical} and \ref{fig:gender-historical-posneg} for racial and gender bias WEATs respectively.

In both the gender and racial temporal analyses, it is clear that the biases previously observed in the embeddings trained on the full corpus of judicial opinions were not primarily the result of the inclusion of historical data.
For the racial bias tests, we observed WEAT scores with moderate to large effect sizes indicative of negative racial bias in both the positive/negative legal WEAT and the grant/deny WEAT at all time periods for African American and Hispanic surnames as compared to European surnames.  The bias effect sizes decreased slightly as less historical data was included for the positive/negative legal attribute WEATs, but remained relatively constant in the grant/deny WEAT.   
For Asian and Pacific Islander surnames as compared to European surnames, we observed the same pattern of negative biases that decrease slightly over time in the positive/negative legal WEAT, but only some time periods were observed to have a moderate negative bias on the grant/deny WEAT, with the strongest observed bias being in the embedding trained on opinions from 2000-2020.

Similar to results from the full word embeddings, the bias scores for gender bias legal WEATs were dependent on the set of targets used, with bias scores near neutral for the generic male/female terms and large bias scores (both positive and negative) observed for the given name-based WEATs. 
While the temporal trends were relatively stable for the generic male/female terms, observed bias scores fluctuated in the given name based measures. 
This may indicate changes in gender bias over time that swing between positive and negative, but it should be noted that the gender specificity of given names also changes over time, making these results more difficult to interpret.

\begin{figure}[tb]
    \centering
    \includegraphics[scale=.33]{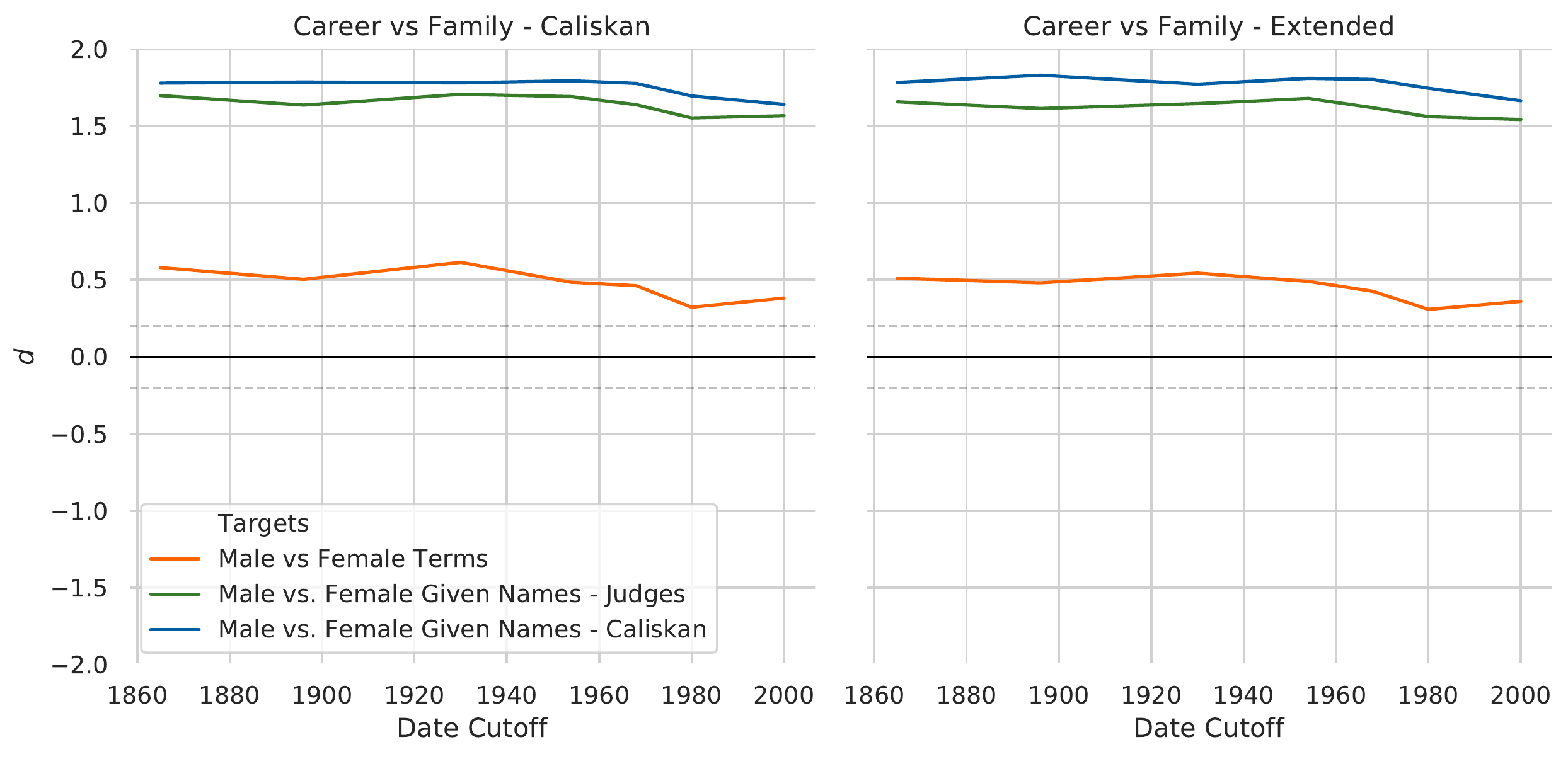}
    \caption{Temporal WEAT scores for gender-related targets and career/family attributes}
    \label{fig:gender-historical-career}
\end{figure}

Since the gender-career bias was the strongest observed in our replication of the original WEAT (see Table~\ref{tab:baseline}), we also performed a temporal analysis of gender-career bias using both the original career and family terms from Caliskan et al. and the expanded set previously described.
Figure~\ref{fig:gender-historical-career} shows that the gender career bias was observed at all time periods and across all gender target types.
This effect was extremely strong for the given name based measures and moderately strong for the male/female terms.

\subsection{Topical Effects} \label{sec:topical}
\begin{figure}[tb]
    \centering
    \includegraphics[scale=.37]{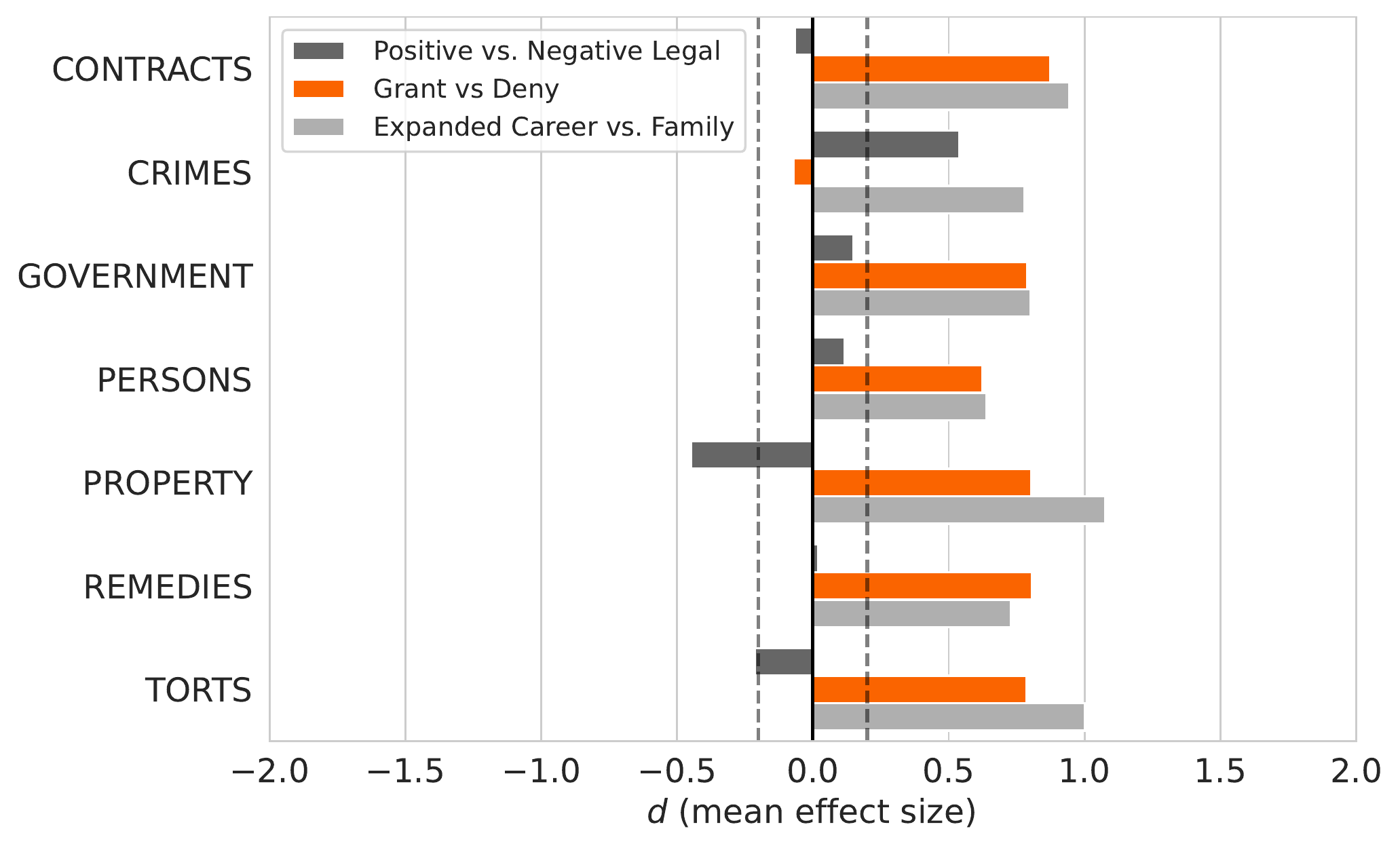}
    \caption{WEAT Cohen's effect sizes for "Male vs. Female Terms" target list. Different attribute lists are shown in different colors.}
    \label{fig:topical-gender}
\end{figure}

In this section, we further investigate how gender and racial biases change when we only consider cases pertaining to a specific legal topic (for results related to racial biases, see the Supplementary Material\footref{fn-supp-mat}). 
To categorize the documents in our dataset, we rely on the seven main divisions of law provided by "West's Analysis of American Law" guide \cite{west-analysis-2013}. 
We define topical areas for each opinion using the Key Number classification for the headnotes written for the opinion. 
The seven main categories are contracts, crimes, government, persons, property, remedies, and torts.
This guide also provides more granular sub-divisions of the main topics, however, for our experiments here we only focus on the main divisions to capture the overall effects observed under each category. 
To make sure the analyses are not affected by a small sample size, while preparing the dataset for each legal category we removed the words in the target and attribute lists that have a frequency of less than 30 occurrences in the corresponding sub-corpus (see "Embedding Training" in "Proposed Approach").
Figures~\ref{fig:topical-gender}~\textendash~\ref{fig:topical-judgename} illustrate the results of some of these tests (see 
the Supplementary Material\footref{fn-supp-mat} for the results of more experiments).

Figure~\ref{fig:topical-gender} shows the results of three tests where the target list is "Male vs. Female Terms". 
We observe that the breakdown of the documents by their legal topic in the case of "Positive vs. Negative Legal" attribute list reveals strong biases in two categories: crimes and property. 
On the other hand, in the case of "Grant vs. Deny" attribute list we observe a significant bias in all legal topics except for crimes. 
Finally, as mentioned in the previous results, there exists significant bias in the case of "Expanded Career vs. Family" attribute list (see similar results for the "Career vs. Family (Caliskan)" attribute list in the Supplementary Material\footref{fn-supp-mat}) and this bias is consistent in terms of the large magnitude across all the different legal topics.

Figures~\ref{fig:topical-name}~and~\ref{fig:topical-judgename} illustrate the results of four tests to compare the detected gender bias for the "Male vs. Female (Caliskan)" attribute lists (Figure~\ref{fig:topical-name}) as a baseline against the legally adapted "Male vs. Female Judge Given Name" lists (Figure~\ref{fig:topical-judgename}). 
These results demonstrate that choosing the legally adapted target lists reveal different type (i.e., sign of the effect size) and magnitude of the bias for each legal topic. 
Observe, for example, that in the case of "Positive vs. Negative Legal" the magnitude of the effect size of the legally adapted lists is smaller compared to the baseline for topics such as property and remedies, and larger in other topics such as crimes, persons, and torts.
We also observe a consistently larger effect size in the case of "Grant vs. Deny" for the legally adapted lists compared to the baseline. 
\begin{figure}[tb]
    \centering
    \includegraphics[scale=.37]{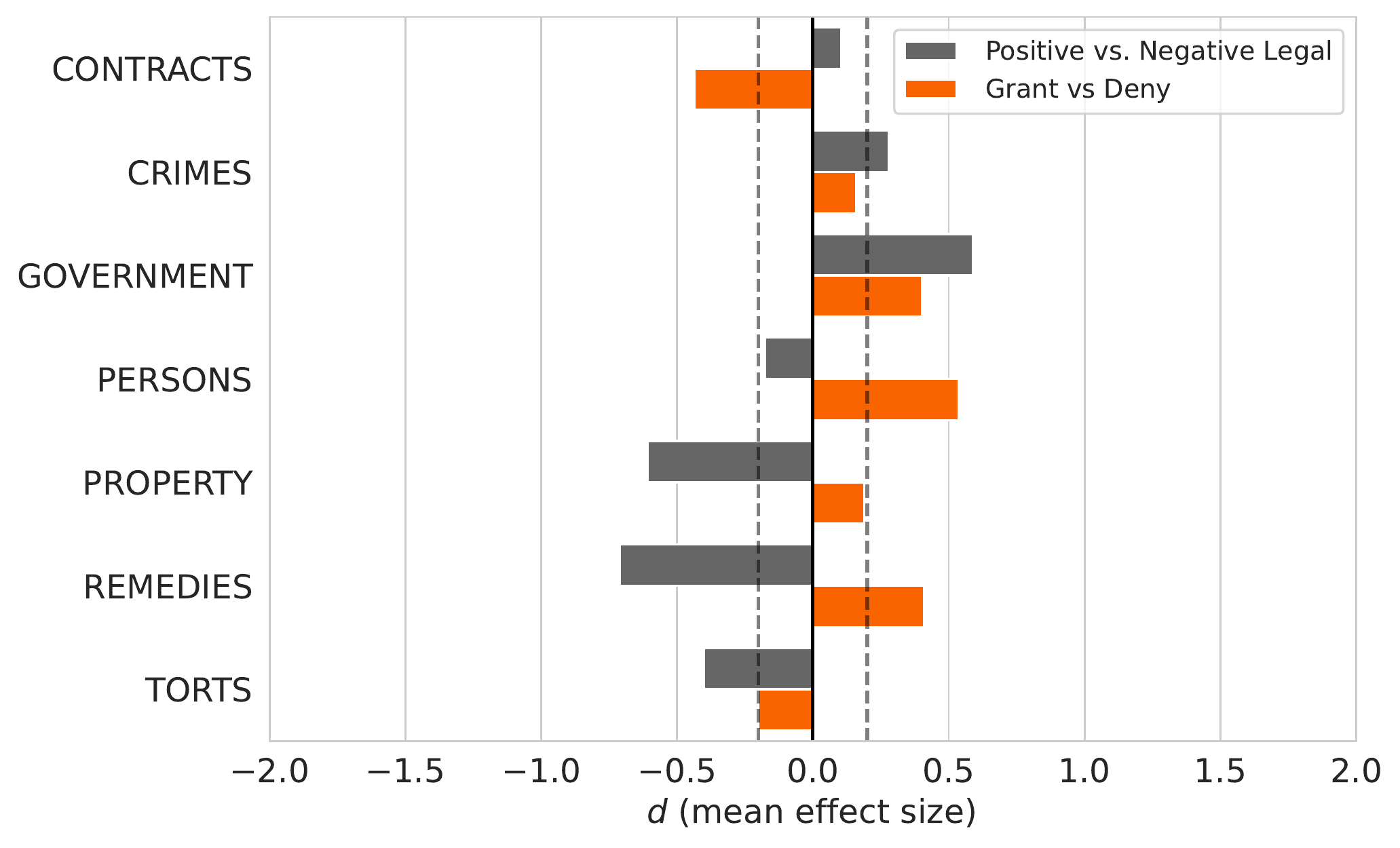} 
    \caption{WEAT Cohen's effect sizes for "Male vs. Female (Caliskan)" target list. Different attribute lists are shown in different colors.}
    \label{fig:topical-name}
\end{figure}
\begin{figure}[tb]
    \centering
    \includegraphics[scale=.37]{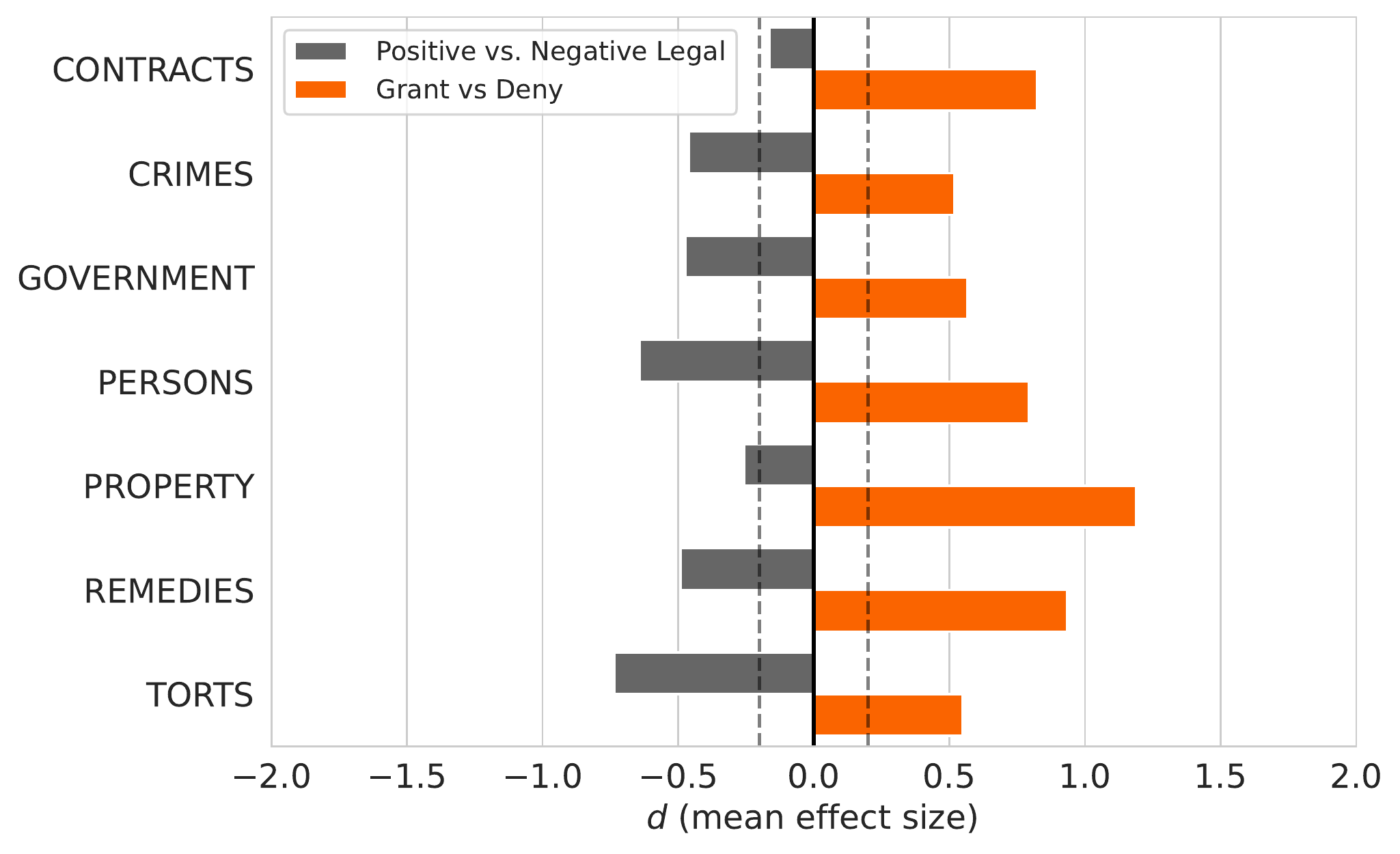} 
    \caption{WEAT Cohen's effect sizes for "Male vs. Female Judge Given Name" target list. Different attribute lists are shown in different colors.}
    \label{fig:topical-judgename}
\end{figure}

\section{Conclusions and Discussion}
In this article we proposed a legally adapted approach for identifying gender and racial biases that are encoded in the word embeddings trained on the text of legal opinions from U.S. case law. 
This approach considers specific idioms used in legal language and also adapts the general bias detection WEAT method to legal language. 
The experiments designed in this work demonstrate the importance of domain adaptation for bias detection methods.
If general purpose bias identification methods are used to measure gender and racial biases in word embeddings in the legal domain or other domains with specialized vocabularies, the developers of these systems may inadvertently create NLP systems that replicate or even amplify these biases in the world even after trying to screen their word embeddings for potential biases.

Using domain adapted bias detection methods is also important for evaluating the effectiveness of any potential mitigation strategy.
We showed that using a date cut off is not an effective strategy for mitigating gender or racial biases present in the legal opinions even though societal opinions regarding these issues have changed over time.
Our results also demonstrate that gender-career bias is particularly strong for given names in this domain, suggesting that downstream legal NLP systems that operate on these representations (e.g., coreference resolution) may be particularly likely to make biased predictions.
Furthermore, we showed that analyzing the bias across different legal topics not only reveals different types of bias but also signifies the need for evaluating the system for fairness under different topics. 

Future work in this area should also focus on the downstream effects exhibited by predictive systems that take biased representations as input as well as the effects any mitigation strategies have on these predictions.
This work examines only biases in the representations themselves but the way that these biases could potentially cause harm in society is when they are used to make predictions that may be biased and the results of these predictions are displayed to users.
The exact nature of the potential harms caused would depend on the specific application, but biased predictions made by these systems could be particularly harmful in contexts where users are not directly viewing the text these models are trained upon but instead are viewing aggregated predictions or summaries of results across many cases.

For example, if a motion outcome prediction system operating on racially biased word representations was deployed within a particularly diverse jurisdiction, it could under count the number of successful motions as compared to model performance in a less diverse jurisdiction.
Attorneys representing clients in this jurisdiction might then be less likely to believe that a potential motion in a client's case would succeed based on a summary of historical outcomes and could suggest settlement in scenarios where they would have proposed continuing with the motion if the model had provided a more accurate prediction of outcomes within their jurisdiction.
Under-representation of counter-stereotypical scenarios in legal research systems due to biases in predictive models operating on biased representations could ultimately contribute to degradation in the quality of legal representation or increased costs related to additional time required for legal research for individuals in protected classes.


\section*{Ethical Statement}
This paper leveraged identity characteristics from the U.S. Census and a judicial biographical database to create target lists for the WEAT test.  This work examined group fairness for both race and gender in word embeddings built from judicial opinions. While the aim of this work is to measure these potentially harmful representational biases in order to facilitate the creation of mitigation strategies for legal NLP systems that take these types of representations as input, the work could also be used to build intentionally harmful or biased legal NLP tools. Unfortunately, blindness itself leads to unfairness and we need to better understand the impact of stereotypes on legal decisions made by the judiciary \cite{nielsen_practical_2020}.

\section*{Acknowledgments}
We would like to thank Frank Schilder, Brian Romer, and Nadja Herger for their guidance and support.


\label{sec:reference_examples}
\bibliography{references}

\begin{thebibliography}{21}
\providecommand{\natexlab}[1]{#1}

\bibitem[{Bolukbasi et~al.(2016)Bolukbasi, Chang, Zou, Saligrama, and
  Kalai}]{bolukbasi_man_2016}
Bolukbasi, T.; Chang, K.-W.; Zou, J.; Saligrama, V.; and Kalai, A. 2016.
\newblock Man is to computer programmer as woman is to homemaker? debiasing
  word embeddings.
\newblock In \emph{Proceedings of the 30th {International} {Conference} on
  {Neural} {Information} {Processing} {Systems}}, {NIPS}'16, 4356--4364. Red
  Hook, NY, USA: Curran Associates Inc.
\newblock ISBN 978-1-5108-3881-9.

\bibitem[{Bouma(2009)}]{bouma_normalized_2009}
Bouma, G. 2009.
\newblock Normalized (pointwise) mutual information in collocation extraction.
\newblock In \emph{Proceedings of the {GSCL}}, 31--40. Tübingen, Germany:
  Gunter Narr Verlag.

\bibitem[{Caliskan, Bryson, and Narayanan(2017)}]{caliskan_semantics_2017}
Caliskan, A.; Bryson, J.~J.; and Narayanan, A. 2017.
\newblock Semantics derived automatically from language corpora contain
  human-like biases.
\newblock \emph{Science (New York, N.Y.)}, 356(6334): 183--186.

\bibitem[{Dieterich, Mendoza, and Brennan(2016)}]{dieterich2016compas}
Dieterich, W.; Mendoza, C.; and Brennan, T. 2016.
\newblock COMPAS risk scales: Demonstrating accuracy equity and predictive
  parity.
\newblock \emph{Northpoint Inc}, 7(7.4): 1.

\bibitem[{{Federal Judicial Center}(2012)}]{fjcexport}
{Federal Judicial Center}. 2012.
\newblock Biographical {Directory} of {Article} {III} {Federal} {Judges}:
  {Export}.
\newblock
  \url{https://www.fjc.gov/history/judges/biographical-directory-article-iii-federal-judges-export}.
\newblock Accessed: 2021-09-03.

\bibitem[{{Free Law Project}(2021)}]{judge_db}
{Free Law Project}. 2021.
\newblock Court Listener: Bulk Judicial Database Files.
\newblock \url{https://www.courtlistener.com/api/bulk-data/people/all.tar.gz}.
\newblock Accessed: 2021-05-03.

\bibitem[{Greenwald, McGhee, and Schwartz(1998)}]{greenwald1998measuring}
Greenwald, A.~G.; McGhee, D.~E.; and Schwartz, J.~L. 1998.
\newblock Measuring individual differences in implicit cognition: the implicit
  association test.
\newblock \emph{Journal of personality and social psychology}, 74(6): 1464.

\bibitem[{Kallus, Mao, and Zhou(2020)}]{kallus_assessing_2020}
Kallus, N.; Mao, X.; and Zhou, A. 2020.
\newblock Assessing algorithmic fairness with unobserved protected class using
  data combination.
\newblock In \emph{Proceedings of the 2020 {Conference} on {Fairness},
  {Accountability}, and {Transparency}}, {FAT}* '20, 110. New York, NY, USA:
  Association for Computing Machinery.
\newblock ISBN 978-1-4503-6936-7.

\bibitem[{Lambrecht and Tucker(2019)}]{lambrecht2019algorithmic}
Lambrecht, A.; and Tucker, C. 2019.
\newblock Algorithmic bias? an empirical study of apparent gender-based
  discrimination in the display of stem career ads.
\newblock \emph{Management Science}, 65(7): 2966--2981.

\bibitem[{Mikolov et~al.(2013{\natexlab{a}})Mikolov, Chen, Corrado, and
  Dean}]{mikolov2013efficient}
Mikolov, T.; Chen, K.; Corrado, G.; and Dean, J. 2013{\natexlab{a}}.
\newblock Efficient Estimation of Word Representations in Vector Space.
\newblock arXiv:1301.3781.

\bibitem[{Mikolov et~al.(2013{\natexlab{b}})Mikolov, Sutskever, Chen, Corrado,
  and Dean}]{mikolov2013distributed}
Mikolov, T.; Sutskever, I.; Chen, K.; Corrado, G.; and Dean, J.
  2013{\natexlab{b}}.
\newblock Distributed Representations of Words and Phrases and their
  Compositionality.
\newblock arXiv:1310.4546.

\bibitem[{Mozafari, Farahbakhsh, and Crespi(2020)}]{mozafari2020hatespeech}
Mozafari, M.; Farahbakhsh, R.; and Crespi, N. 2020.
\newblock Hate speech detection and racial bias mitigation in social media
  based on BERT model.
\newblock \emph{PLOS ONE}, 15(8): 1--26.

\bibitem[{Nielsen(2020)}]{nielsen_practical_2020}
Nielsen, A. 2020.
\newblock \emph{Practical {Fairness}: {Achieving} {Fair} and {Secure} {Data}
  {Models}}.
\newblock O'Reilly Media, Incorporated.
\newblock ISBN 978-1-4920-7573-8.

\bibitem[{Obermeyer et~al.(2019)Obermeyer, Powers, Vogeli, and
  Mullainathan}]{obermeyer2019dissecting}
Obermeyer, Z.; Powers, B.; Vogeli, C.; and Mullainathan, S. 2019.
\newblock Dissecting racial bias in an algorithm used to manage the health of
  populations.
\newblock \emph{Science}, 366(6464): 447--453.

\bibitem[{Pennington, Socher, and Manning(2014)}]{pennington-etal-2014-glove}
Pennington, J.; Socher, R.; and Manning, C. 2014.
\newblock {G}lo{V}e: Global Vectors for Word Representation.
\newblock In \emph{Proceedings of the 2014 Conference on Empirical Methods in
  Natural Language Processing ({EMNLP})}, 1532--1543. Doha, Qatar: Association
  for Computational Linguistics.

\bibitem[{Rice, Rhodes, and Nteta(2019)}]{rice_racial_2019}
Rice, D.; Rhodes, J.~H.; and Nteta, T. 2019.
\newblock Racial bias in legal language.
\newblock \emph{Research \& Politics}, 6(2): 2053168019848930.
\newblock Publisher: SAGE Publications Ltd.

\bibitem[{Rice and Zorn(2019)}]{dvn_2019}
Rice, D.; and Zorn, C. 2019.
\newblock {Replication Data for: "Corpus-Based Dictionaries for Sentiment
  Analysis of Specialized Vocabularies"}.
\newblock Harvard Dataverse, V1, \url{https://doi.org/10.7910/DVN/4EKHFM}.

\bibitem[{Rice and Zorn(2021)}]{rice_corpus-based_2021}
Rice, D.~R.; and Zorn, C. 2021.
\newblock Corpus-based dictionaries for sentiment analysis of specialized
  vocabularies.
\newblock \emph{Political Science Research and Methods}, 9(1): 20--35.

\bibitem[{Suresh and Guttag(2020)}]{suresh2020framework}
Suresh, H.; and Guttag, J.~V. 2020.
\newblock A Framework for Understanding Unintended Consequences of Machine
  Learning.
\newblock arXiv:1901.10002.

\bibitem[{{Thomson Reuters Westlaw}(2013)}]{west-analysis-2013}
{Thomson Reuters Westlaw}, ed. 2013.
\newblock \emph{West's Analysis of American Law}.
\newblock Westlaw.

\bibitem[{Vacek et~al.(2019)Vacek, Song, Molina-Salgado, Teo, Cowling, and
  Schilder}]{vacek-etal-2019-litigation}
Vacek, T.; Song, D.; Molina-Salgado, H.; Teo, R.; Cowling, C.; and Schilder, F.
  2019.
\newblock Litigation Analytics: Extracting and querying motions and orders from
  {US} federal courts.
\newblock In \emph{Proceedings of the 2019 Conference of the North {A}merican
  Chapter of the Association for Computational Linguistics (Demonstrations)},
  116--121. Minneapolis, Minnesota: Association for Computational Linguistics.

\end{thebibliography}

\newpage
\appendix
\section*{Supplementary Material}
\section{Derived Target and Attribute Word Lists}
Here we provide the derived target and attribute lists used for the main experiments. See \citeauthor{caliskan_semantics_2017} \citeyearpar{caliskan_semantics_2017} for the baseline target and attribute lists.


\subsection{Positive vs. Negative Legal}

These attribute lists were derived from \citeauthor{rice_corpus-based_2021} \citeyearpar{rice_corpus-based_2021} seed term queries against the Legal Opinion Corpus embeddings.  The seed terms are as follows:

\begin{itemize}
    \item \textbf{Positive Seeds:} agree, correct, favorable, perfect, intelligent, agreement, faithful, wise, consistent, convinced
    \item \textbf{Negative Seeds:} bad, wrong, mistaken, fail, mistake, unnecessary, unsupported, untenable, reject, erroneous
\end{itemize}

The corresponding query results as well as the terms excluded after manual review are as follows:

\subsubsection{Positive Legal:}
    
    faithful, guaranteeing, amiable, pleasant, cheerful, appreciative, personable, businesslike, courteous, healthful, congenial, strived, undertakings, dependable, loving, faithfully, agreeable, thanks, cordial, honorable, splendid, grateful, energetic, harmonious, respectful, loyal, rapport, assuring, bodied, vibrant, wonderful, fullest, desirous, striven, obedient, admired, nurturing, cheerfully, enthusiastically, decorous, compliment, affable, guarantying, generations, commodious, fulfillment, sociable, guarantied, intelligent, punctual, finest, wholesome, amplest, unhindered, decent, contemporaries, considerate, hearty, likeable, punctually, dignified, proud, easygoing, industrious, charming, cordially, prosperous, affectionate, friendly, polite, thrifty, pleasantly, fairest, sacredly, keen, fulfilment, prided, economical, wonderfully, delighted, engagements, aspire, decently, talented, minutest, praised, pluralistic, superintend, excellency, dutiful, insightful, proudly, lively, beautifully, hopeful, excelled

\subsubsection{Positive Legal Excluded:}
    manly, gentlemanly, Bachelor
        
\subsubsection{Negative Legal:}
    
    unsupported, incorrect, erroneous, wrong, irrelevant, unfounded, untenable, baseless, improper, erroneously, mistaken, mistake, meritless, inadequate, arbitrary, bad, immaterial, premised, faulty, insufficient, improperly, ignored, disregarded, inadmissible, patently, groundless, unavailing, impermissible, inaccurate, fail, reject, misplaced, excessive, unnecessary, flawed, unjustified, inappropriate, capricious, incorrectly, unsubstantiated, vague, manifestly, predicated, inapplicable, unwarranted, unsupportable, unfair, unmeritorious, invalid, misdirected, unreasonable, lacking, untimely, unreliable, unconstitutional, impermissibly, rejecting, unrelated, speculation, rejection, misleading, specious, justification, confusing, unauthorized, incomplete, misapplied, unjust, misstates, manifest, frivolous, unpersuasive, reversible, merit, fallacious, abandoned, justified, justify, duplicative, undue, irrational, mischaracterization, illegal, grossly, misreading, arbitrarily, mistakenly, recklessly, rejects, disregard, premature, insufficiency, incredible, futile, subjective, wrongdoing, plainly, ignoring, motive, irrelevancy

\subsection{Legal (Motion) Outcome}

The following attribute lists were derived from disposition terms for legal motions and appeals:

\subsubsection{Grant:}
    
    grant, grants, granting, granted, accept, accepts, accepted, accepting, affirm, affirms, affirmed, affirming, approve, approves, approved, approving, sustain, sustains, sustained, sustaining
    
\subsubsection{Deny:}
    
    deny, denies, denying, denied, decline, declines, declined, declining, vacate, vacates, vacated, vacating, decline, declines, declined, declining, overrule, overrules, overruled, overruling

\subsection{Expanded Career vs. Family}

The attribute lists were derived from \citeauthor{rice_corpus-based_2021} \citeyearpar{rice_corpus-based_2021} seed term queries against the Legal Opinion Corpus embeddings.   The seed terms from WEAT 6 in \citeauthor{caliskan_semantics_2017} \citeyearpar{caliskan_semantics_2017} are as follows:

\begin{itemize}
    \item \textbf{Career Seeds:} executive, management, professional, corporation, salary, office, business, career.
    \item \textbf{Family Seeds:} home, parents, children, family, cousins, marriage, wedding, relatives.
\end{itemize}

The query results as well as the terms excluded after manual review are as follows:

\subsubsection{Expanded Career:}
    
    executive, chief-executive, managerial, salaried, vice-president, salary, operations, operational, Chief-Executive, corporate, CEO, director, COO, president, CFO, management, revenue, board-of-directors, Corporate, chairman, Chief-Financial, hiring-and-firing, organizational-structure, Vice-President, salaries, senior-vice, managing, President, Executive-Vice, corporation, Executive, Chief-Operating, personnel, payroll, Marketing, executives, fiscal, Board-of-Directors, subsidiary, functions, regulatory, automation, duties, managers, commissions, delegated, clerical, assistant, marketing, employing, comptroller, Senior-Vice, vice-presidents, Managing-Director, entity, oversight, audit, Delaware-corporation, departments, usurping, supervisory, President-and-CEO, executive-branch, wholly-owned-subsidiary, offices, Operations, engineering, affiliate, private-sector, performs, demoting, directors, professional, shareholder, competitive, company, reorganizing, Professional, usurpation, secretary-treasurer, revenues, annual-salary, engineer, manager, human-resource, directorship, foreign-corporation, internal, profitability, Comptroller, operating, engineers, Management, promotion, forecasting, restructuring, auditors, competitor, branch, policymaking
    
\subsubsection{Expanded Family:}
    
    cousins, grandparents, aunts, grandmother, stepmother, aunt, aunt-and-uncle, paternal-grandparents, siblings, mother, maternal-grandfather, maternal-grandparents, sisters, maternal-grandmother, stepfather, children, paternal-grandmother, paternal-grandfather, stepchildren, uncles, daughters, maternal-relatives, godmother, maternal-uncle, relatives, daughter, granddaughters, maternal-aunt, paternal-aunt, paternal, parents, grandchildren, youngest, eldest, uncle, cousin, grandchild, granddaughter, niece, twins, nieces, younger-brother, father, youngest-child, younger-siblings, fiance, Aunt, grandson, brothers-and-sisters, stepbrother, boyfriend, childrens, estranged, sister, fiancee, Grandmother, grandsons, older-siblings, nephews, mothers, stepsister, minor-children, nieces-and-nephews, grandfather, biological-parents, boyfriends, reunited, stepson, maternal, paramour, foster-parents, son, Grandparents, adoptive, teenage, sons, Daughter, stepdaughters, minor-child, girlfriend, biological-father, married, stepdaughter, loved, nephew, adoptive-parents, friends, out-of-wedlock, childless, girls, wedding, kin, girlfriends, loving, teenagers, teenaged, roommates, mom-and-dad

\subsubsection{Expanded Family Excluded:}
    Brianna, Tabitha

\subsection{Surnames by Race}

The following Surname lists were sampled from the 2010 US Census.  As noted in the paper, each name was required to appear in at least 300 opinions except for the Native American and Alaskan Native list. 

\subsubsection{European Last Names:}
    
Rae, Crisco, Deeley, Bjorklund, Mardian, Aloi, Loewen, Schuster, Engelmann, Ulery, Fiorenzo, Buis, Haycock, Hickory, Hudelson, Lembke, Milbauer, Heffelfinger, Gribble, Mahler, Balestra, Lutz, Brincat, Kandel, Dileo, Marter, Frymire, Nielson, Sirota, Callison, Boydston, Yeager, Dressler, Sachs, Guhl, Hufnagle, Warshaw, Spodek, Saporito, Hegel, Borgeson, Hogland, Balick, Rinke, Stunkard, Hegstrom, Donahey, Mastronardi, Zweig, Kleban, Ineichen, Tunheim, Gudgel, Rhomberg, Rohrs, Henslee, Lobdell, Prins, Mohr, Gillson, Simoneaux, Wetherington, Avers, Paine, Piesco, Serota, Hottenstein, Moskowitz, Torkelson, Solly, Scovel, Goerke, Nemecek, Scruton, Montesano, Dekker, Dray, Toomey, Lamson, Caffrey, Pingree, Milos, Offill, Kralik, Roley, Grabowski, Inglese, Barstad, Sestito, Estey, Englander, Hirshfield, Karibian, Secor, Arrowood, Ludtke, Wenner, Silverberg, Klinck, Coxon, Raborn, Kraushaar, Creely, Pellerito, Kiker, Tallman, Spath, Slee, Riedlinger, Lisle, Kleinschmidt, Abbatiello, Zagar, Farquhar, Hudlow, Aulicino, Verni, Caney, Latona, Leif, Sommers, Melear, Hardt, Filippo, Ollis, Cassano, Giaccio, Rosenman, Longval, Moeser, Agosti, Malony, Sayer, Caswell, Borowsky, Steffey, Dreyer, Thorman, Halferty, Fridley, Berwald, Tyndall, Formby, Famolare, Winkle, Devall, Severtson, Cloutier, Brindley, Betz, Leonardi, Goetzman, Kraemer, Fronk, Trafford, Setter, Giuliano, Guilmette, Conkwright, Ramstad, Cathell, Sundheim, Ebert, Vigliotti
    
\subsubsection{African American Last Names:}
    
Pettaway, Jessamy, Ephriam, Sinkfield, Senegal, Pondexter, Minnifield, Bendolph, Osagie, Okeke, Boateng, Okoro, Mensah, Cephas, Claybrooks, Vaughns, Hardnett, Cephus, Whack, Ndiaye, Kennebrew, Owusu, Madyun, Bangura, Acoff, Hameen, Chukwu, Conteh, Malveaux, Philmore, Dumpson, Marbley, Ojo, Golphin, Mems, Mercadel, Akande, Narcisse, Knowlin, Wigfall, Lavalais, Sinegal, Lucious, Gaitor, Hargro, Idowu, Torain, Tresvant, Adeniji, Eleby, Bluitt, Luvene, Broaden, Opoku, Addo, Lawal, Shabazz, Ajayi, Bloodsaw, Grandberry, Roulhac, Bodison, Asante, Ducksworth, Killings, Honora, Glasper, Twymon, Poullard, Adu, Hypolite, Whitsey, Beyah, Adeleke, Wrice, Madu, Glinsey, Teamer, Earvin, Wrighten, Broadnax, Sails, Nwachukwu, Gadsden, Cudjoe, Jubilee, Osei, Taiwo, Smalls, Wimes, Salaam, Gadsen, Batiste, Prioleau, Chatmon, Anyanwu, Stepney, Woodfolk, Okafor, Blige, Menefield, Tukes, Okoli, Adeyemi, Lately, Tolefree, Geathers, Presha, Arvie, Fluellen, Ofori, Bacote, Seabrooks, Outing, Wysinger, Manigault, Diallo, Expose, Yeboah, Gabbidon, Baymon, Balogun, Haynesworth, Snype, Ancrum, Nutall, Pinkins, Peguese, Okoye, Boykins, Aytch, Ravenell, Hugee, Afriyie, Shelvin, Darensburg, Winbush, Veasley, Macharia, Straughter, Villery, Tasby, Hezekiah, Neverson, Blakes, Petties, Yelder, Contee, Holiness, Goffney, Degrate, Akpan, Junious, Leffall, Stanciel, Jiggetts, Dunkins, Gadson, Summage, Smokes, Cooperwood, Poitier, Eze, Taybron
    
\subsubsection{Hispanic Last Names:}
    
Montes, Ocana, Sanabria, Magallon, Bejarano, Camarillo, Fierros, Oviedo, Guevara, Melendrez, Becerril, Osorio, Reynoso, Villasenor, Zepeda, Gastelo, Zacarias, Pomales, Montelongo, Galeana, Mazariegos, Abrego, Garfias, Palacios, Zorrilla, Oquendo, Recinos, Alderete, Iraheta, Zurita, Delgadillo, Aleman, Saldivar, Mendieta, Miramontes, Tellez, Inzunza, Escobar, Cuadrado, Beltre, Penaloza, Coreas, Cardena, Villalba, Rubalcaba, Rizo, Taveras, Echeverry, Medina, Batres, Vences, Carmona, Matamoros, Lazcano, Bencomo, Lizarraga, Alvarenga, Costilla, Preciado, Segovia, Villeda, Aparicio, Yanez, Callejas, Salinas, Estrada, Pulido, Botello, Magdaleno, Cobian, Govea, Medellin, Escobedo, Nava, Conejo, Reynosa, Ascencio, Guajardo, Cardona, Saenz, Santoyo, Galvan, Baez, Adames, Benitez, Sauceda, Cerda, Loaiza, Veliz, Zamorano, Valadez, Pelayo, Veloz, Navarrete, Manjarrez, Polanco, Basurto, Herrera, Espinoza, Obeso, Arzola, Sarabia, Perdomo, Rubio, Ovalle, Arrellano, Rengifo, Jasso, Lombera, Pantoja, Cobos, Sosa, Sanchez, Berroa, Montalvo, Mejia, Alcala, Huerta, Chavez, Solorzano, Olmedo, Morfin, Bastidas, Terrones, Alverio, Giraldo, Mayorga, Lagunas, Lozoya, Olivas, Aguilar, Placencia, Brizuela, Calvillo, Rosalio, Guebara, Carrasco, Germosen, Urias, Olivarez, Cervantes, Zamudio, Banales, Liranzo, Ibarra, Flores, Alvarez, Gonzalez, Cedillo, Altamirano, Galaviz, Villagra, Barrientos, Campuzano, Zuniga, Robledo, Yepez, Cadena, Vargas, Ovando, Genao, Hermosillo, Alatorre, Morales, Mireles, Lemus, Nogueras, Posada, Tapia, Aldaco, Oropeza, Fragozo, Puerta, Pizano, Beniquez, Astorga, Jerez, Reynaga, Rivas, Ortega, Murillo, Colmenares, Limon, Amezquita, Chairez, Mariscal, Abarca, Nuno, Ortuno, Carranza, Aceves, Rincon, Zamora, Mosqueda, Cornejo, Arciniega, Retana, Camarena, Londono, Tamez
    
\subsubsection{Asian Pacific Island Last Names:}
    
Chung, Saxena, Doshi, Tsui, Vue, Mehta, Ryu, Lu, Luk, Lui, Hyun, Nam, Sung, Wang, Chan, Yi, Chon, Liew, Lieu, Shih, Tan, Tuan, Hsieh, Huang, Panchal, Parekh, Parikh, Ravi, Bhatt, Hoang, Ma, Diep, Nghiem, Shukla, Kwok, Tian, Kao, Jia, Mao, Pathak, Lor, Thi, Bae, Manoharan, Rajesh, Shin, Jeong, Yim, Satish, Huong, Rong, Hou, Qian, Choi, Ou, Cheng, Vitug, Giang, Hu, Moua, Tse, Iyer, Kwon, Garg, Lai, Luong, Saechao, Quach, Kuang, Jie, Thanh, Ip, Suh, Guo, Vuong, Dinh, Li, Tam, Pak, Ng, Le, Truong, Huynh, Bui, Chuang, Duong, Gautam, Thakkar, Cao, Vu, Ho, Vang, Leang, Yuan, Mui, Song, Ye, Eun, Agrawal, Ha, Keung, Desai, Yum, Kang, Chae, Chiang, Bansal, Shen, Teng, Szeto, Dao, Phong, Jin, Cho, Ding, Agarwal, Sundaram, Kwan, Kulkarni, Aggarwal, Shu, Bhatnagar, Thao, Jang, Sanghera, Hwa, Phu, Shetty, Hsu, Srinivasan, Gandhi, Tsai, Wei, Thang, Yue, Leung, Yin, Choe, Kuo, Poon, Gu, Naik, Chui, Tsang, Tang, Deng, Pei, Chen, Wen, Nguyen, Bhakta, Chuan, Chua, Hwang, Saelee, Phan, Zhou, Han, Tsao, Chu, Xu, Tseng, Vo, Chih, Luu, Su, Nanavati, Goyal, Pham, Kyung, Patel, Trinh, Chau, Zhang, Yu, Kyong, Gupta, Fu, Won, Dang, Goswami, Trivedi, Cai, Ly, Kothari, Trung, Yun, Khurana, Zhao, Adusumilli, Liang, Vyas, Seung, Xiong, Seo, Xiao, Zhu, Liao, Yeh, Pandya
    
\subsubsection{Native American Last Names:}
    
Whiteface, Madplume, Stillday, Cheromiah, Denetclaw, Blackbear, Yellowhair, Begaye, Tsosie, Etsitty, Yepa, Greyeyes, Youngbird, Cowboy, Manygoats, Neztsosie, Quiver, Yazzie, Halona, Calabaza, Blackhorse, Whiteplume, Youngbear, Manuelito, Peshlakai, Haskie, Atcitty, Becenti, Spoonhunter, Peneaux, Kingbird, Benally, Bluebird, Tsinnijinnie, Wassillie, Nez, Hosteen, Kameroff, Zunie, Ganadonegro, Laughing, Chischilly, Fasthorse, Wauneka, Bedonie, Goldtooth    

\subsection{Male vs. Female Terms}

The following gendered pronouns and common nouns were used for the gender based tests.

\subsubsection{Male Terms:}
    
male, males, man, men, boy, boys, he, him, his, himself
    
\subsubsection{Female Terms:}
    
female, females, woman, women, girl, girls, she, her, hers, herself

\subsection{Judge Given Names}

The following given name lists were sampled from a Judicial biographical database \cite{fjcexport}. 

\subsubsection{Male Judge Given Name:}
    
    Dennis, Joe, Howard, Stanley, Daniel, Anthony, Bernard, William, Harold, Raymond, Kenneth, Samuel, Carl, Brian, Sidney, Roger, Alfred, Horace, Vincent, Eric, Douglas, David, Richard, Larry, Andrew, Herbert, Benjamin, Steven, Walter, Warren, Timothy, Charles, Tom, Jon, Kevin, Maurice, Allen, Earl, Henry, Terry, Matthew, Jerry, Gregory, Leonard, Arthur, Frank, Fred, Ralph, Edwin, James, Sam, Jeffrey, Scott, Robert, George, Harry, Alexander, Albert, Gary, John, Ernest, Mark, Jesse, Peter, Clarence, Eugene, Joseph, Marvin, Hugh, Michael, Francis, Donald, Nicholas, Stephen, Paul, Christopher
    
\subsubsection{Female Judge Given Name:}
    
    Alice, Amy, Ann, Anna, Anne, Barbara, Beth, Brenda, Carmen, Carol, Carolyn, Catherine, Cathy, Cheryl, Christine, Cynthia, Deborah, Debra, Denise, Diana, Diane, Donna, Elizabeth, Ellen, Helen, Holly, Jacqueline, Jane, Janet, Janice, Jennifer, Jill, Joan, Judith, Julie, Karen, Katherine, Kathleen, Kimberly, Laura, Laurie, Linda, Lisa, Lori, Louise, Marcia, Margaret, Maria, Marilyn, Marsha, Martha, Mary, Maureen, Michelle, Nancy, Pamela, Patricia, Paula, Phyllis, Rebecca, Robin, Rosemary, Ruth, Sandra, Sara, Sarah, Sharon, Shirley, Stephanie, Sue, Susan, Suzanne, Teresa, Vanessa, Victoria, Wendy

\section{Temporal Effects: More Experiments}
Here we provide the results of more experiments in our temporal study. 

\begin{figure}[tb]
    \centering
    \includegraphics[scale=.4]{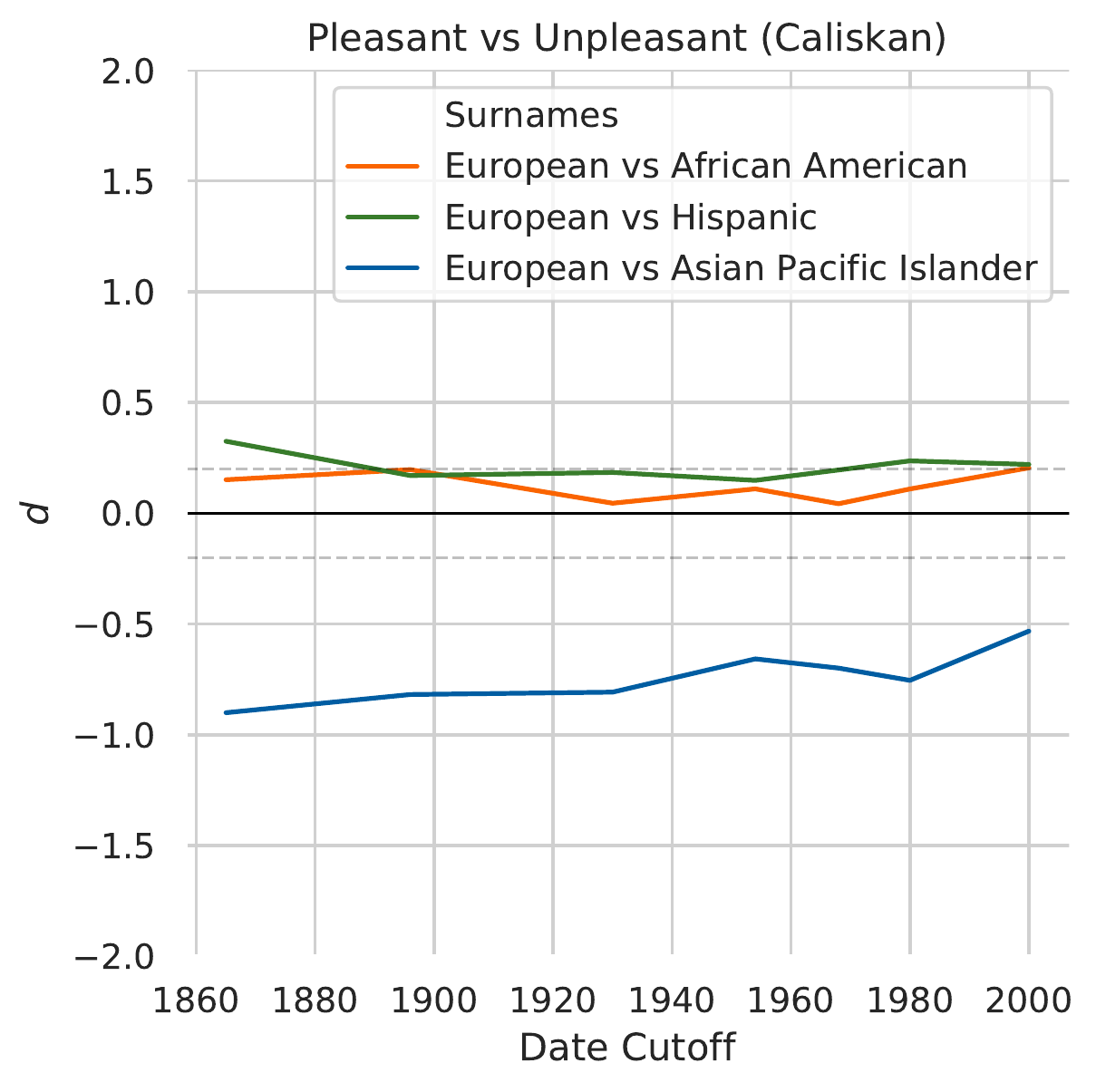}
    \caption{Temporal WEAT scores for race targets (surnames) and pleasant/unpleasant attributes from \citeauthor{caliskan_semantics_2017} \citeyearpar{caliskan_semantics_2017}}
    \label{fig:temporal-appendix-race}
\end{figure}

\begin{figure}[tb]
    \centering
    \includegraphics[scale=.4]{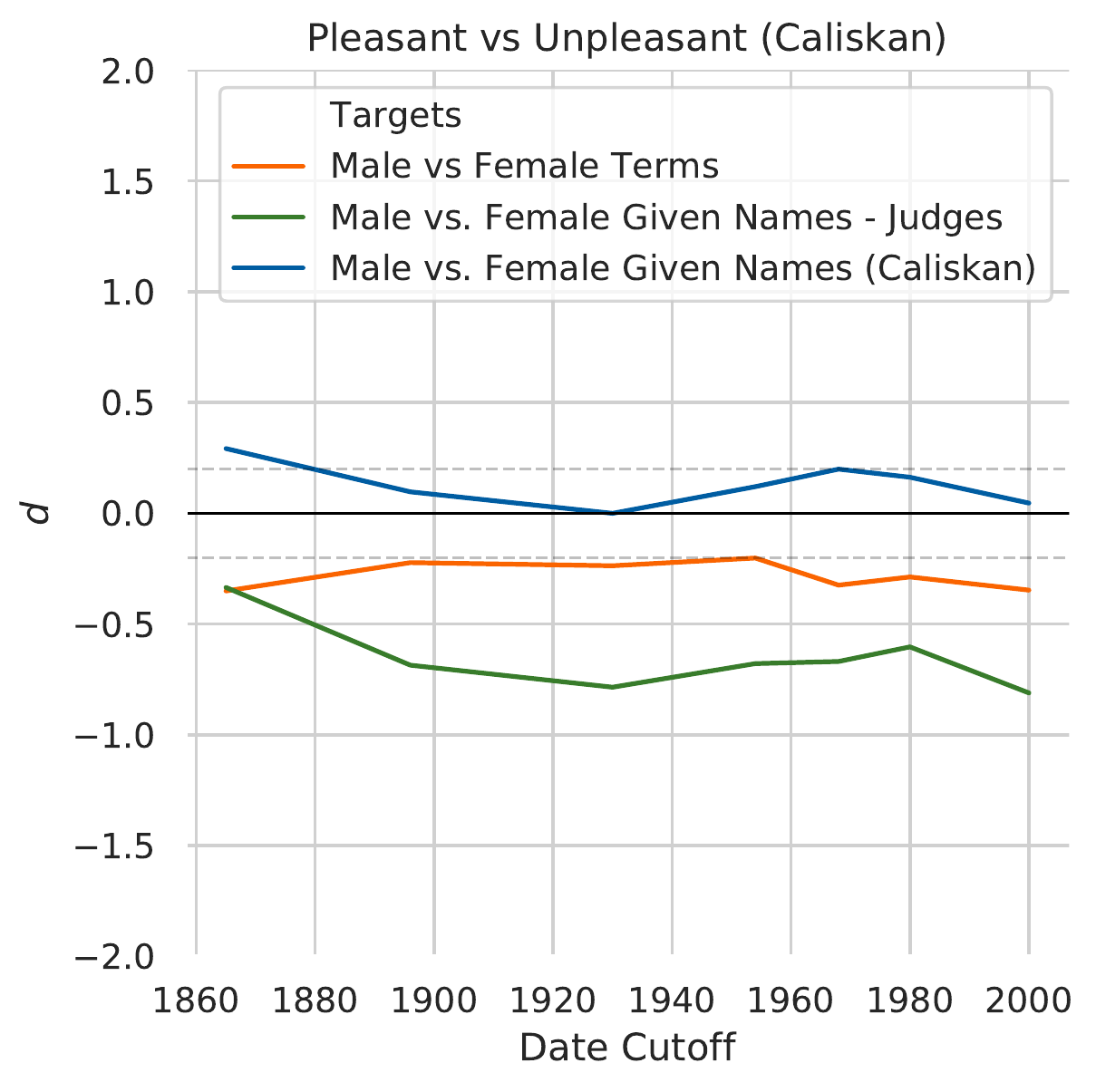}
    \caption{Temporal WEAT scores for gender-related targets and pleasant/unpleasant attributes from \citeauthor{caliskan_semantics_2017} \citeyearpar{caliskan_semantics_2017}}
    \label{fig:temporal-appendix-gender}
\end{figure}

Figures~\ref{fig:temporal-appendix-race} and ~\ref{fig:temporal-appendix-gender} show the results of temporal WEATs using the \citeauthor{caliskan_semantics_2017}\citeyearpar{caliskan_semantics_2017} pleasant/unpleasant attributes with race surname and gender targets resepectively.

\section{Topical Effects: More Experiments}
Here we provide the results of more experiments in our topical study. 

\begin{figure*}[tb]
    \centering
    \includegraphics[scale=.4]{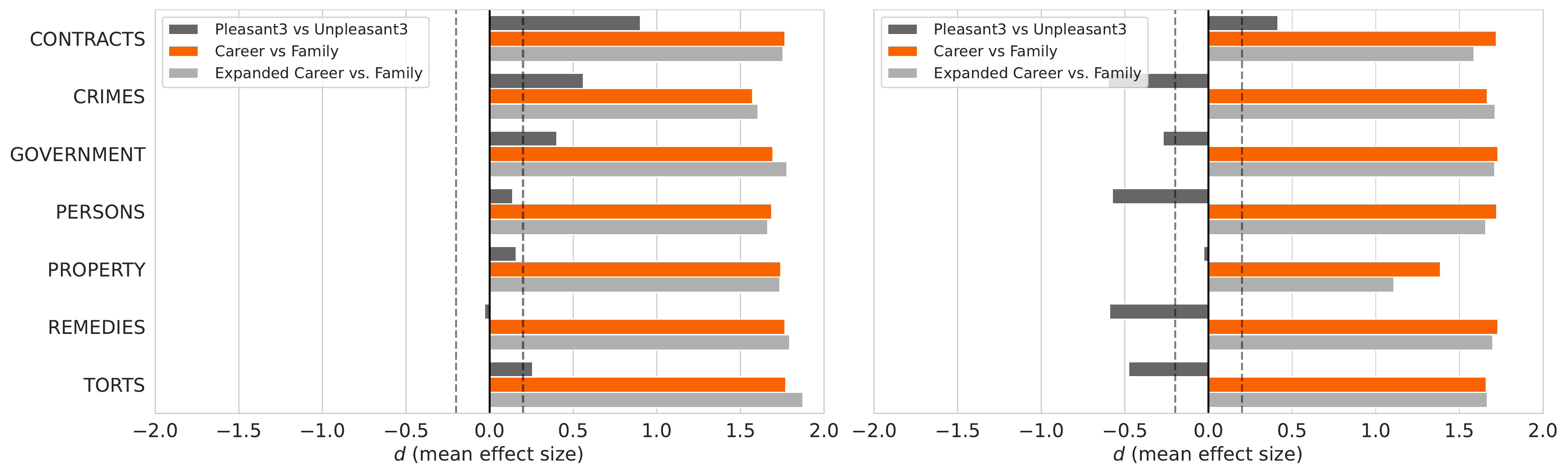}
    \caption{WEAT Cohen's effect sizes for "Male vs. Female (Caliskan)" target list on the left and "Male vs. Female Judge Given Name" target list on the right. Different attribute lists (i.e., "Pleasant3 vs Unpleasant", "Career vs. Family", and "Expanded Career vs. Family") are shown in different colors.}
    \label{fig:topical-appendix-judgename}
\end{figure*}

Figure~\ref{fig:topical-appendix-judgename} illustrates the results of six tests to compare the detected gender bias for the "Male vs. Female (Caliskan)" attribute lists as a baseline against the legally adapted "Male vs. Female Judge Given Name" lists which complement the results provided in section~"Topical Effects". 

\begin{figure}[tb]
    \centering
    \includegraphics[scale=.4]{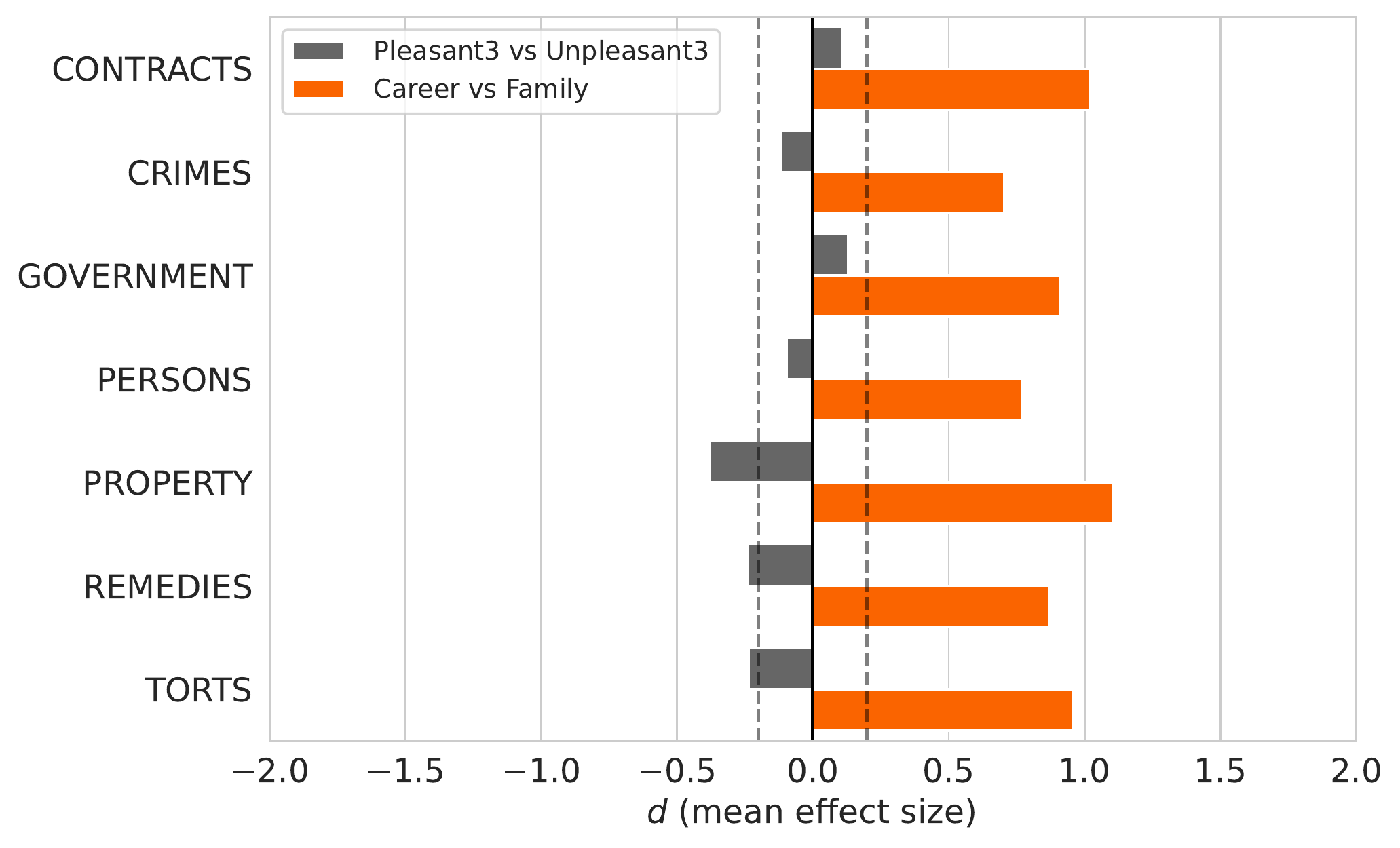}
    \caption{WEAT Cohen's effect sizes for "Male vs. Female Terms" target list. Different attribute lists (i.e., "Pleasant3 vs. Unpleasant3" and "Career vs. Family") are shown in different colors.}
    \label{fig:topical-appendix-gender}
\end{figure}

Figure~\ref{fig:topical-appendix-gender} shows the results of two tests where the target list is "Male vs. Female Terms". 
We observe that the breakdown of the documents by their legal topic in the case of "Pleasant3 vs. Unpleasant3" (from Caliskan) attribute list reveals strong bias only in one category: property, and all the legal topics show a strong bias in the case of "Career vs. Family" (from Caliskan) similar to the expanded list provided in section~"Topical Effects". 

\begin{figure}[hbt!]
    \centering
    \includegraphics[scale=.4]{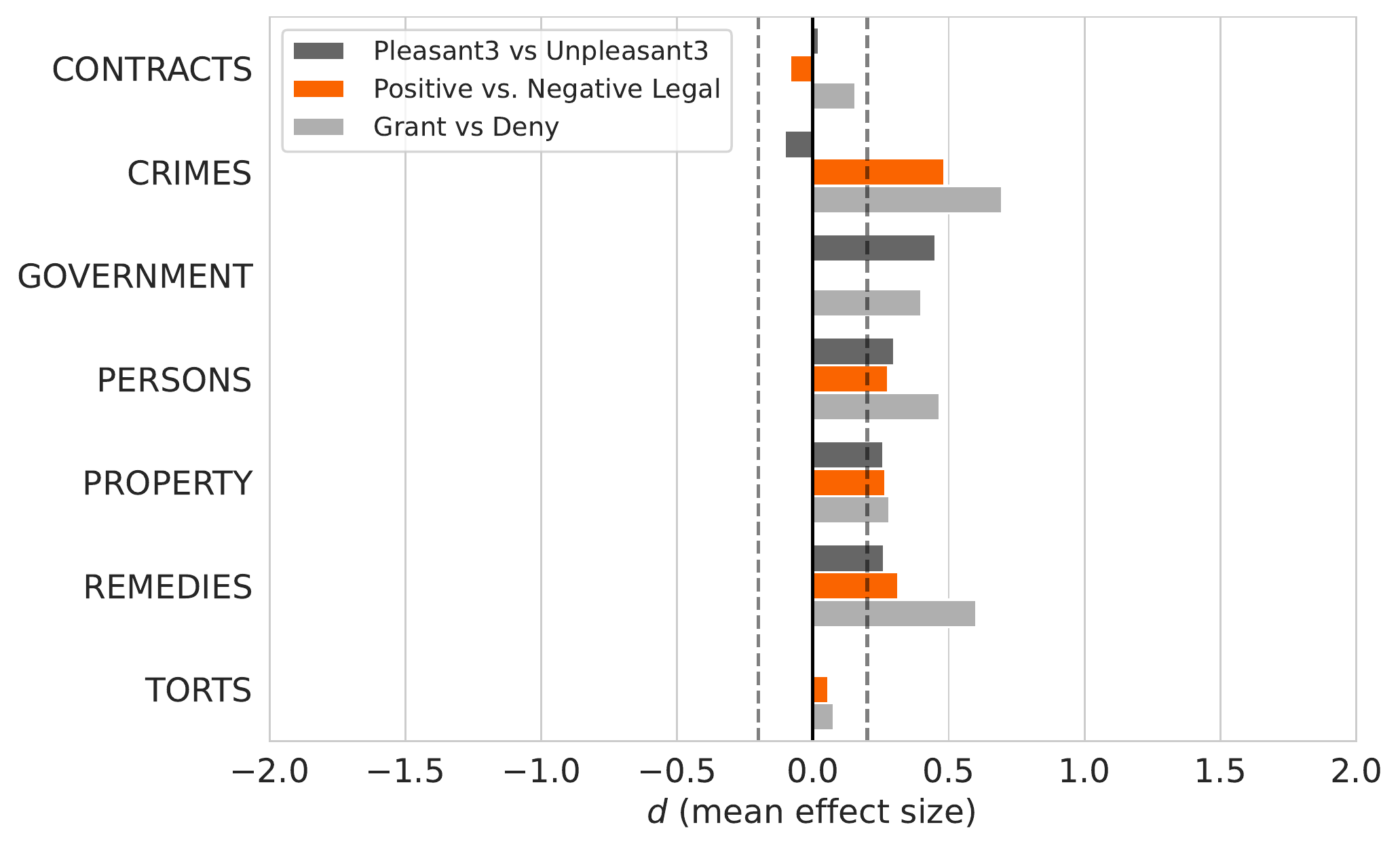}
    \caption{WEAT Cohen's effect sizes for "European vs. African American Last Names" target list. Different attribute lists are shown in different colors.}
    \label{fig:topical-appendix-lastname_EUAA}
\end{figure}

\begin{figure}[hbt!]
    \centering
    \includegraphics[scale=.4]{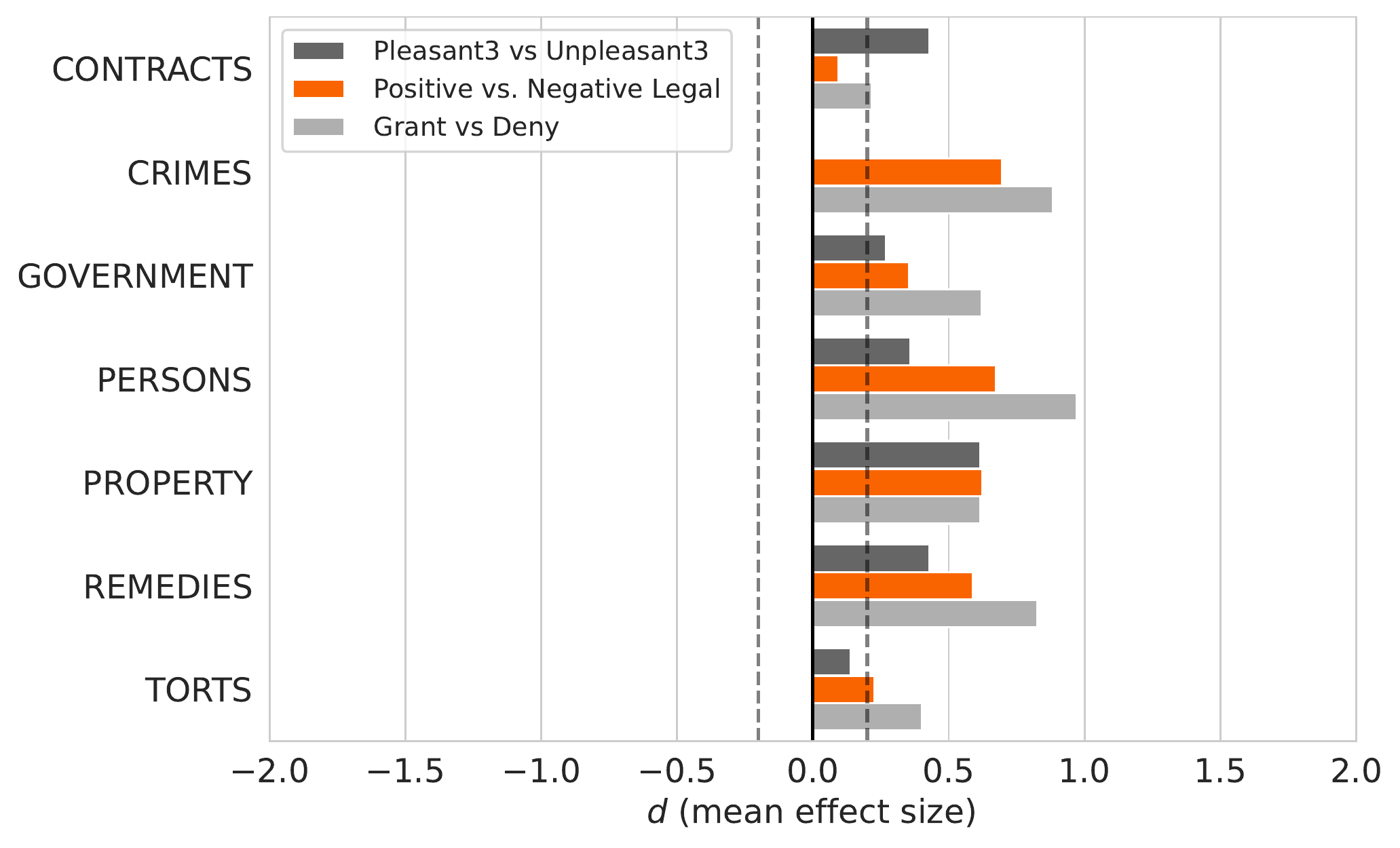}
    \caption{WEAT Cohen's effect sizes for "European vs. Hispanic Last Names" target list. Different attribute lists are shown in different colors.}
    \label{fig:topical-appendix-lastname_EUH}
\end{figure}

\begin{figure}[hbt!]
    \centering
    \includegraphics[scale=.4]{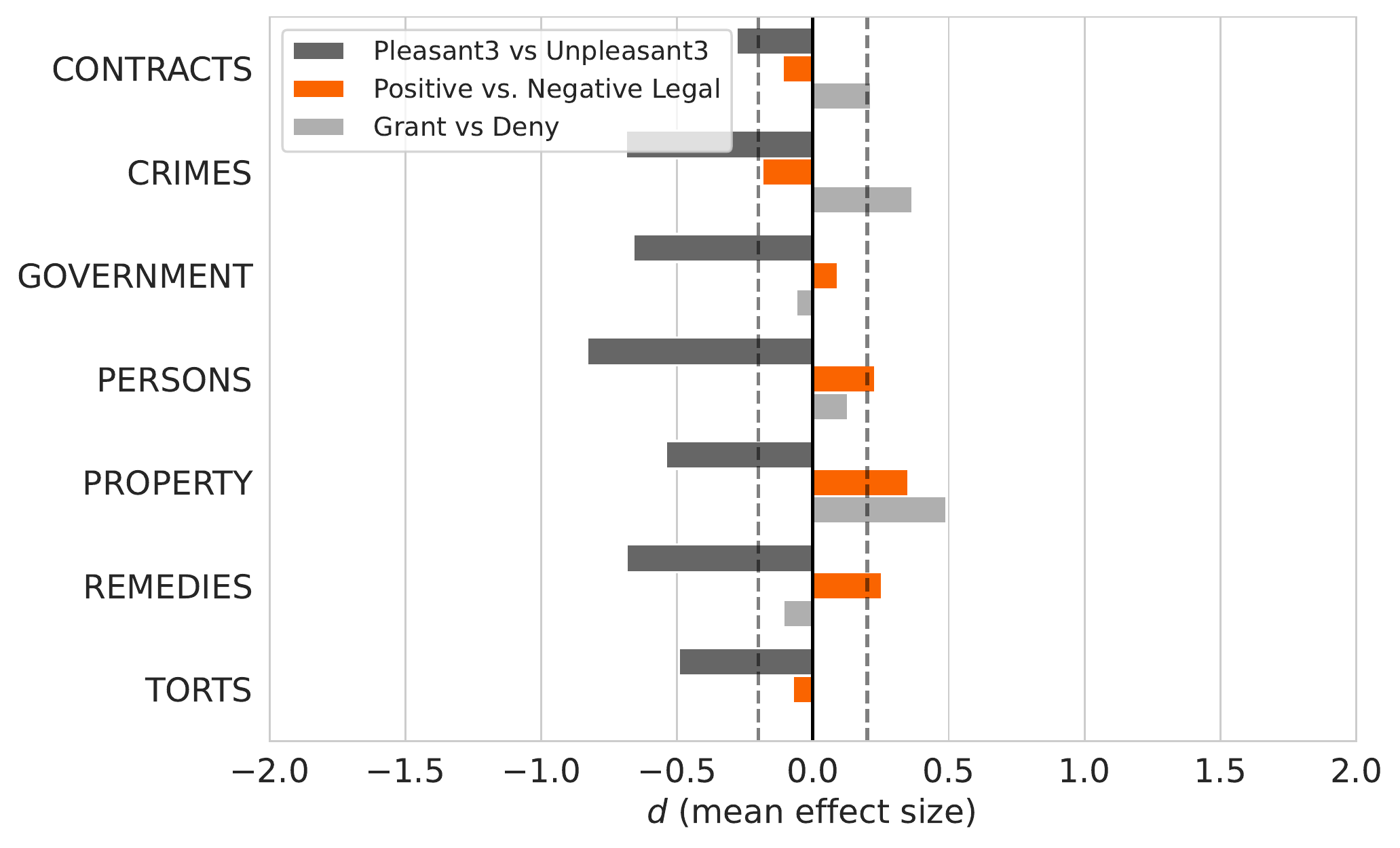}
    \caption{WEAT Cohen's effect sizes for "European vs. Asian Pacific Island Last Names" target list. Different attribute lists are shown in different colors.}
    \label{fig:topical-appendix-lastname_EUAPI}
\end{figure}

\begin{figure}[hbt!]
    \centering
    \includegraphics[scale=.4]{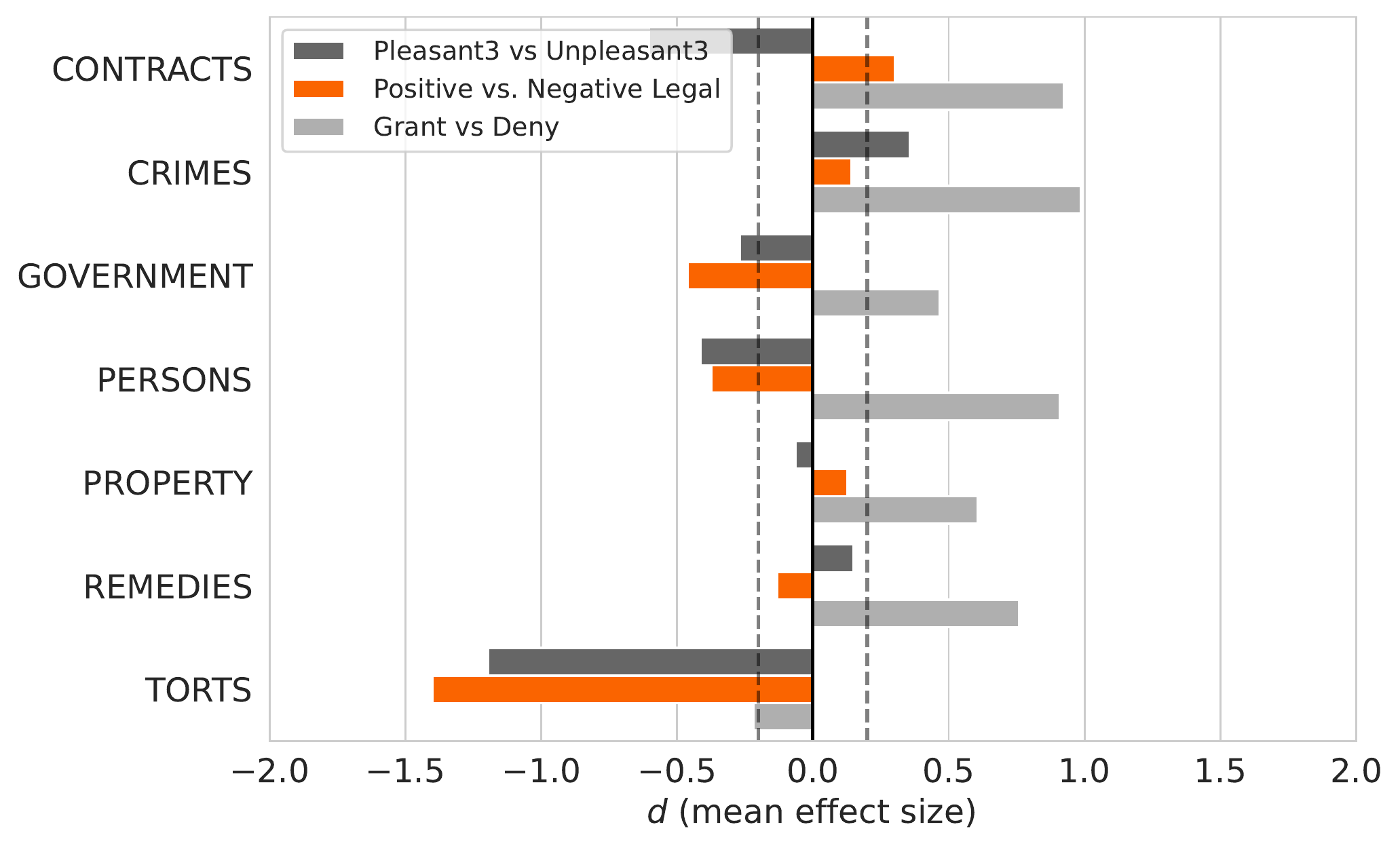}
    \caption{WEAT Cohen's effect sizes for "European vs. Native American Last Names" target list. Different attribute lists are shown in different colors.}
    \label{fig:topical-appendix-lastname_EUNA}
\end{figure}

Figures~\ref{fig:topical-appendix-lastname_EUAA}-\ref{fig:topical-appendix-lastname_EUNA} illustrate the results of WEAT tests on different racial last name target lists. 
These figures reveal different types and magnitudes of detected bias across racial groups and topics of law. 
\newpage

\newpage

\newpage

\end{document}